\RequirePackage{fix-cm}
\documentclass[twocolumn]{svjour3}          % twocolumn
\smartqed  % flush right qed marks, e.g. at end of proof
\usepackage{graphicx}
\usepackage{latexsym}
\usepackage[caption=false]{subfig}
\usepackage[ruled,vlined]{algorithm2e}
\usepackage{amssymb}
\usepackage{ragged2e}
\usepackage{makecell}
\usepackage{multirow}
\usepackage{amsmath}
\usepackage{mathtools}
\usepackage{hyperref}
\usepackage{tabularx}
\usepackage{color,soul}
\usepackage{cite}
\usepackage{calc}

\newlength{\firsttermlength}

\usepackage{amsmath}
\begin{document}

\title{Toward Real-Time Image Annotation Using Marginalized Coupled Dictionary Learning\thanks{Authors contributed equally on this research.}%\thanks{Grants or other notes
}
\subtitle{}


\author{Seyed Mahdi Roostaiyan		\and
		Mohammad Mehdi Hosseini		\and
		Mahya Mohammadi Kashani		\and
		S. Hamid Amiri        
}

\authorrunning{Roostaiyan et al.} % if too long for running head

\institute{Seyed Mahdi Roostaiyan \at
              Department of Computer Engineering, Sharif University of Technology, Azadi Ave., Theran, Iran \\
              \email{mahdiroostaiyan@ce.sharif.edu}           %  \\
           \and
           Mohammad Mehdi Hosseini \at
              Department of Computer Engineering, Sharif University of Technology, Azadi Ave., Theran, Iran \\
              \email{mohammadmehdi.hosseini@du.edu}
           \and
           Mahya Mohammadi Kashani \at
              Department of Computer Engineering, Shahid Rajaee Teacher Training University, Lavizan, Theran, Iran \\
              \email{mahya.mkashani@sru.ac.ir}
           \and
           S. Hamid Amiri \at
              Department of Computer Engineering, Shahid Rajaee Teacher Training University, Lavizan, Theran, Iran \\
              \email{s.hamidamiri@sru.ac.ir}
}


\maketitle

\begin{abstract}
In most image retrieval systems, images include various high-level semantics, called tags or annotations. Virtually all the state-of-the-art image annotation methods that handle imbalanced labeling are search-based techniques which are time-consuming. In this paper, a novel coupled dictionary learning approach is proposed to learn a limited number of visual prototypes and their corresponding semantics simultaneously. This approach leads to a real-time image annotation procedure. Another contribution of this paper is that utilizes a marginalized loss function instead of the squared loss function that is inappropriate for image annotation with imbalanced labels. We have employed a marginalized loss function in our method to leverage a simple and effective method of prototype updating. Meanwhile, we have introduced ${\ell}_1$ regularization on semantic prototypes to preserve the sparse and imbalanced nature of labels in learned semantic prototypes. Finally, comprehensive experimental results on various datasets demonstrate the efficiency of the proposed method for image annotation tasks in terms of accuracy and time. The reference implementation is publicly available on \href{https://github.com/hamid-amiri/MCDL-Image-Annotation}{\textcolor{blue}{github}}.

%\subclass{MSC code1 \and MSC code2 \and more}
\end{abstract}

%%%%%%%%%%%%%%%%%%%%%%%%%%%%%% INTRODUCTION %%%%%%%%%%%%%%%%%%%%%%%%%%%%%
\section{Introduction}
\label{sec_intro} 
\begin{figure}
\includegraphics[width=0.48\textwidth]{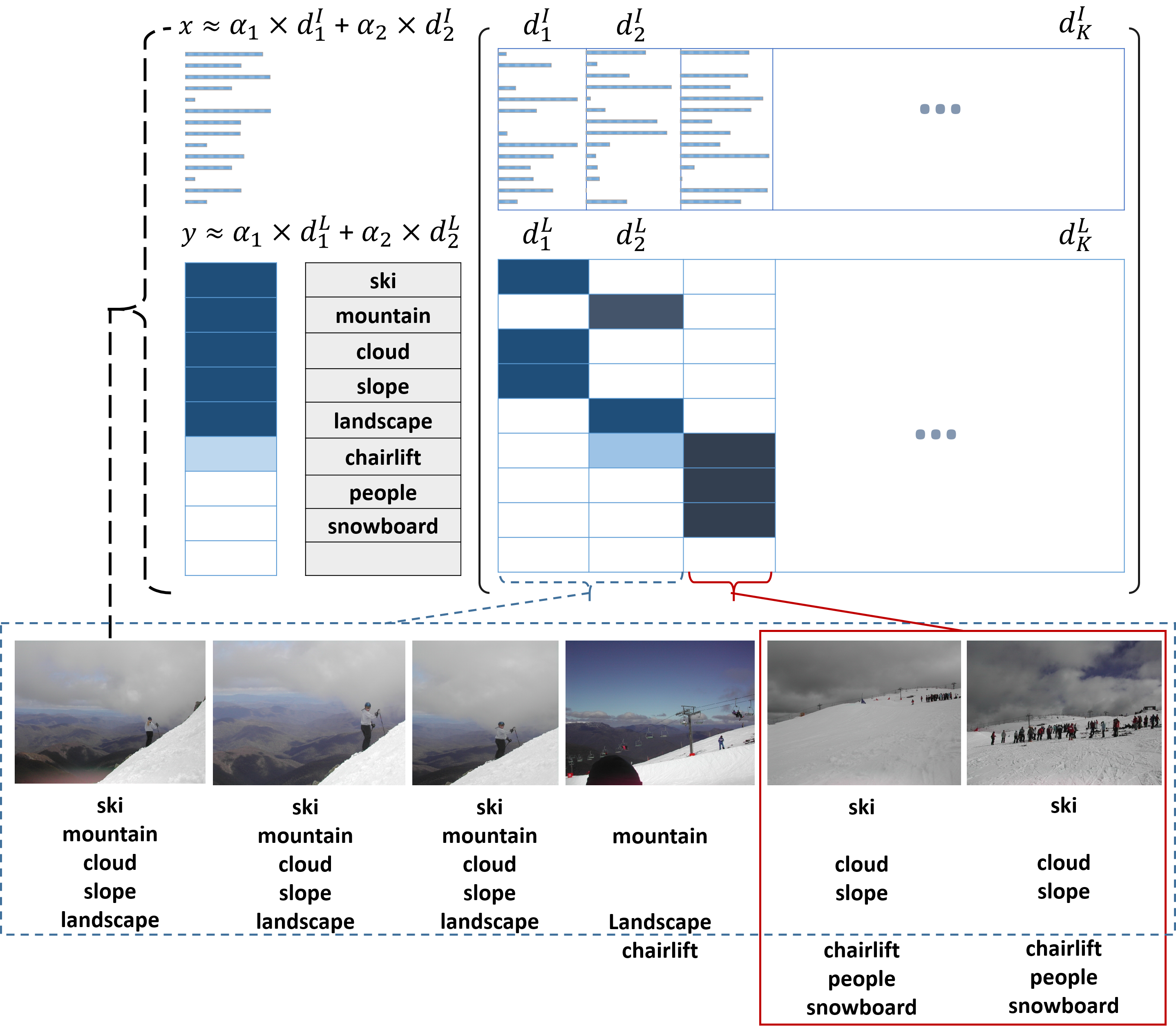}
\caption{Abstract of data summarization using MCDL. Example images have been taken from IAPRTC-12 dataset. There are shared visual and semantic contents in these images that can be summarized into three representatives.}
\label{fig_summarization}       % Give a unique label
\end{figure}

Image annotation deals with the problem in which each instance is represented by a single example that is associated with multiple labels. The main challenges of the image annotation problem, which distinguishes it from standard multi-label problems,  are “class-imbalance” (extreme variations in the frequency of different labels), “incomplete-labeling” (many images are not annotated with all the relevant labels of the vocabulary) \cite{verma2017image}, and “diverse-labeling” (predicted labels must be qialified representative of the image and diverse from each other, to reduce redundancy) \cite{verma2019diverse}. Since the early approaches of image annotation (e.g., generative-based models \cite{putthividhy2010topic}) did not consider these challenges, they have low performance in the annotation task. \\
Similarity-based strategy \cite{guillaumin2009tagprop, verma2017image} is arguably the most intuitive solution that annotates a given image based on its nearest neighbors, which is an effective approach for image annotation tasks concerning the aforementioned challenges. Considering the potential weakness of this strategy, which ignores correlations between labels, various approaches such as metric learning \cite{verma2017image, bragantini2021rethinking} and sparse multi-view multi-label learning \cite{zhang2020multi} focused on both the visual contents of images and their corresponding labels, simultaneously. Metric learning \cite{verma2017image} aims to learn an improved similarity measure to enhance the efficacy of nearest-neighbor based approaches. However, it is a time-consuming task to compare a query image with all images in the dataset and select the most similar images for annotating a query image. Making a lot of comparisons leads to an unefficient query time in image annotation, so moving toward a real-time image annotation system is a demanded issue in the real-world applications.\\
On the other hand, in large scale datasets, there are some images with similar information in their visual contents and semantic labels. Figure \ref{fig_summarization} shows samples from IAPRTC-12 dataset that are visually and semantically similar. Therefore, it is essential to reduce the redundancy of annotated datasets without missing the original information in the labeled images. To reduce redundancy of the dataset, achieve a real-time annotation mechanism, and presrve the generalization of the annotation method, we follow this strategy in which labeled images could be replaced by a set of representatives, called prototypes in this paper. Figure \ref{fig_summarization} illustrates the idea of data summarization based on the prototype learning. As this figure shows, the primary purpose of the proposed method, called  \mbox{\textit{Marginalized}} \mbox{\textit{Coupled}}  \mbox{\textit{Dictionary}}  \mbox{\textit{Learning}} (MCDL), is to factorize images and their corresponding labels as a weighted sum of learned prototypes. To achieve this goal, MCDL method learns visual (image) prototypes and their corresponding semantic (label) counterparts simultaneously.\\ 

In this paper, we have suggested a joint optimization problem to minimize the reconstruction and hinge losses w.r.t the visual and semantic dictionaries. The discrimination term utilized in MCDL can be regarded as a modification of ${ \ell}_1-norm$ Support Vector Machine \mbox{\cite{zhu20041, hastie2015statistical}}, which was first presented for feature selection in high dimensional feature spaces. The number of prototypes is usually greater than the number of positive samples for each label, which can raise the issue of overfitting. MCDL utilizes ${ \ell}_1$ regularization to learn semantic prototypes as sparse as possible. Another inspiration to use ${ \ell}_1$ regularization is that label vectors are sparse by nature, and each visual prototype could correspond to a few labels. \\
In the literature, similar strategies have been suggested for multi-label classification \mbox{\cite{sun2018sparse, li2019discriminative}}, and image annotation \mbox{\cite{000030-jing16}} to incorporate label discrimination in the dictionary learning stage. Most of these methods utilize squared loss function \mbox{\cite{5539989, 000030-jing16}} for both image and label modalities, aiming to reduce coding residual w.r.t training samples. When using the squared loss function for tags that are naturally imbalanced with many zero entries, label reconstructions are biased to zero (Figure \mbox{\ref{fig_margin}-a}) due to the symmetric property of the squared loss function. This will decrease decision margin and lead to less generalization for the annotation step. To tackle these issues, we have suggested a marginalized loss function with ${\ell}_1$ regularization on semantic prototypes. Taking advantage of the marginalized hinge loss function and ${\ell}_1$ regularization, MCDL could obtain labeled prototypes with admissible generalization in the test stage. To sum up, the main contributions of this paper are as follows:
\begin{itemize}
\item
A coupled dictionary learning strategy is proposed to factorize labeled images into visual prototypes and their corresponding semantic vectors.
\item
MCDL employs the hinge loss for semantic modality, which imposes loss values just for false positives and false negatives. While conventional coupled dictionary learning approaches, such as  Discriminative K-SVD (D-KSVD) \mbox{\cite{5539989}}, Label Consistent K-SVD (LC-KSVD) \mbox{\cite{jiang2011learning}}, and Multi-label Dictionary Learning (MLDL) \mbox{\cite{000030-jing16}}, employ the squared loss defined over binary labels (D-KSVD) or codes (LC-KSVD and MLDL) for label discrimination which impose unnecessary reconstruction loss for true positives and true negatives.
\item
Each visual prototype can be associated with a few semantic tags. We employ ${ \ell}_1$ regularization to impose this sparsity prior knowledge about semantic prototypes. This avoids the overfitting originating from high dimensional space, especially when a few positive samples are available for a label.
\end{itemize}
The rest of this paper is organized as follows. In Section \ref{sec_background}, the related works are reviewed. Section \ref{sec_notations} introduces the definitions and notations used in this paper. The proposed MCDL method and details of the annotation strategy are described in Sections \ref{sec_proposedmethod} and \ref{sec_annotation}. Experimental settings and results are then presented in Section \ref{sec_expRes}. Finally, we conclude our work in Section \ref{sec_conclusion}.

%%%%%%%%%%%%%%%%%%%%%%%%%%%%%%%% NOTATIONS %%%%%%%%%%%%%%%%%%%%%%%%%%%%%%% 
\section{Notations}
\label{sec_notations}
Table \ref{tab_notations} summarizes the notations used in this paper. Suppose that there are $N$ training samples illustrated by $\mathcal{X}=\{(x^1,y^1),\dots,(x^N,y^N)\}$, where the $i^{th}$ image consists of two modalities: 1- the visual modality $x^{i} \in \mathbb{R}^M$ ($M$ is the dimensionality of feature vector), 2- the semantic modality $y^{i} \in {\{0,1\}}^T$ ($T$ is the number of distinct labels). A non-zero entry of $y^{i}$ means that the given image has been annotated by the associated label. The number of annotations for $i^{th}$ sample is denoted by ${{\mathcal{N}}_{+}^i}$. The number of annotated samples with $t^{th}$ label is also denoted by $N_t^{+}$. Moreover, by concatenating different training vectors, we define $\mathbf{X} = \left[x^1,\dots,x^N \right] \in \mathbb{R}^{M\times N}$ and $\mathbf{Y} = \left[y^1,\dots,y^N \right] \in \mathbb{R}^{T\times N}$, which respectively denote image and label data matrices. \\
In our formulation, the coupled dictionary is depicted by ${\mathbf{D}}^C = {\left[ {{\mathbf{D}}^I}^{\top}, {{\mathbf{D}}^L}^{\top} \right]}^{\top} \in \mathbb{R}^{(M+T)\times K}$, where $ M \ll K < N$ is the number of prototypes. This dictionary is composed of two sub-dictionaries ${\mathbf{D}}^I \in \mathbb{R}^{M\times K}$  and ${\mathbf{D}}^L \in \mathbb{R}^{T \times K}$ for visual and semantic modalities respectively. The coefficients matrix is also shown by $\mathbf{A}=[{\alpha}^1,...,{\alpha}^N] \in \mathbb{R}^{K\times N}$, where ${\alpha}^i$ is the sparse representation of $i^{th}$ training sample. The $k^{th}$ column of ${\mathbf{D}}^C$ is also called a coupled prototype denoted as $d_k^C$. Each coupled prototype consists of two sub-prototypes depicted by $d_k^I$ and $d_k^L$ for visual and label modalities. In this paper, ${\mathbf{d}}_{r,k}^L$ denotes the $r^{th}$ row and $k^{th}$ column of the semantic dictionary. To denote the $r^{th}$ row of this matrix we have used ${\mathbf{d}}_r^L = {\mathbf{D}}_{r,.}^{L}$.

\begin{table} 
\caption{Symbols and notations used in this paper.}
\label{tab_notations}      % Give a unique label
\begin{tabular}{r|l}
\noalign{\smallskip}\hline\noalign{\smallskip}
$N$ & The number of training samples \\
$M$ & Dimensionality of feature vectors\\
$T$ & The number of labels (tags)\\
${\mathbf{X}} \in \mathbb{R}^{M\times N}$ & Images matrix \\
${\mathbf{Y}} \in \mathbb{R}^{T\times N}$ & Labels matrix \\
$K$ & Num of learned prototypes\\
${{\mathcal{N}}_{+}^i}$ & Num of annotations for $i^{th}$ sample \\
$N_t^{+}$ & Num of positive samples for $t^{th}$ tag \\
${\mathbf{D}}^I \in \mathbb{R}^{M\times K}$ & Visual dictionary \\
${\mathbf{D}}^L \in \mathbb{R}^{T\times K}$ & Semantic dictionary \\
${\mathbf{D}}^C \in \mathbb{R}^{(M+T)\times K}$ & Coupled dictionary \\
$\mathbf{A}  \in \mathbb{R}^{K\times N}$ & Coefficients matrix \\
$d_k^C \in \mathbb{R}^{(M+T)}$ & $k^{th}$ column of coupled dictionary \\
$d_k^I \in \mathbb{R}^{(M)}$ & $k^{th}$ column of visual dictionary \\
$d_k^L\in \mathbb{R}^{(T)}$ & $k^{th}$ column of semantic dictionary \\
${\mathbf{d}}_t^L \in \mathbb{R}^{1\times K}$ & $t^{th}$ row of semantic dictionary \\
\noalign{\smallskip}\hline
\end{tabular}
\end{table}

%%%%%%%%%%%%%%%%%%%%%%%%%%%% LITERATURE REVIEW %%%%%%%%%%%%%%%%%%%%%%%%%%% 
\section{Literature Review}
\label{sec_background}
Due to the challenges discussed in the previous section, many approaches for handling these challenges belong to the search-based methods \cite{chen2020image, jiang2011learning, verma2019diverse}, by this assumption that the more visual similarity between two images, the more common labels among them. In 2PKNN \cite{verma2017image, verma2019diverse}, the authors proposed a two-pass version of the k-nearest neighbor technique for image annotation. To annotate an image, this method firstly retrieves the most similar images for each label, then computes an image-to-label similarity score, as well as utilizing a metric learning strategy for improving the image-to-image similarity measure.\\
Needing to compute the similarity of a query image to all images in the dataset, search-based methods are inherently time-consuming which emphasizes the importance of introducing scalable methods. Regardless of mete-data-based approaches \cite{sun2020context, samih2021improving}, different methods have been suggested for scalable image annotation, which can be categorized into three main groups, including prototype-based \cite{00009-shooroki17}, dimensionality-reduction-based \cite{heo2018distance, luo2018robust}, and transform-based methods \cite{000030-jing16, zhang2020multi}. The prototype-based approaches cluster samples and then choose one or a few samples or their representatives in each cluster \cite{00009-shooroki17}. Dimensionality-reduction-based approaches, such as product quantization \cite{heo2018distance} and hashing \cite{luo2018robust}, focus on encoding high-dimensional feature spaces densely to achieve speed-up in search-based methods as well as reducing the memory costs. Our proposed approach belongs to the third group of scalable methods, transform-based approaches \cite{000030-jing16}, that treat image annotation as a multi-label problem. In these approaches, both visual and semantic modalities are incorporated into the learning procedure for transforming input data into another space with higher levels of discrimination. One of the successful techniques in this category is sparse representation whose objective is to represent each pattern just using the linear combination of a few numbers of prototypes. Traditional sparse representation approaches can be considered as unsupervised methods that either ignore label information \cite{mairal2010online} or learn prototypes for each label separately. In recent years, many researchers have focused on embedding label information into the prototype learning procedure, generally known as discriminative \cite{li2019discriminative} or coupled \cite{song2018multimodal} dictionary learning, extensively applied for multi-label classification problems \cite{wang2017cross, sun2018sparse}.\\
Discriminative sparse models have many applications in image classification, super-resolution \mbox{\cite{zhao2019image}}, fault-diagnosis, etc. class-specific and shared discriminative dictionary learning (CASDDL) method \mbox{\cite{zhou2021jointly}} aims to classify the steel sheets based on the Fisher discrimination method. They strive to extract the discriminative features for each class separately (inter-class information), along with a shared sub-dictionary which is common between all the classes for extracting the intra-class information. Li et al. \mbox{\cite{li2019discriminativeMCD}} offered a weighted regularization approach to tackle the noisy images. They separate the coarse and fine structures of the noisy images by discriminative sparse methods. Class-oriented discriminative dictionary learning (CODLL) \mbox{\cite{ling2019class}} is another discriminative-based method that not only maximizes the discrimination power of the dictionary atoms, but also considers the discrimination of the coefficients. They limited the atoms to make a group that describes a specific class and simultaneously restricted the coefficients to reconstruct data utilizing the class-related group of atoms.  Structured discriminant analysis dictionary learning \mbox{(SDADL)} \mbox{\cite{du2021structured}}  aims to learn a structured discriminant analysis dictionary. This structured dictionary consists of class-specific sub-dictionaries.  SDADL also introduces a classification loss term to learn and an optimal linear classifier. To continue, we introduce three different sparse-based discriminative methods with more details. \\
Motivated by the success of coupled dictionary learning for classification problems, similar techniques, such as semantic label embedding dictionary (SLED) \cite{cao2015sled}, MLDL \cite{000030-jing16}, and MSFS \cite{zhang2020multi} have been employed for annotation. SLED uses $||\mathbf{X}-\mathbf{D} \mathbf{A}||_F^2+||\mathbf{A}||_1+\Omega(\mathbf{A})$, where $||\mathbf{X}-\mathbf{D}\mathbf{A}||_F^2$ strives to transform the visual training data into a new space, describable with the minimum atoms of matrix $\mathbf{A}$. The sparsity condition is controlled by the second term, i.e. $||\mathbf{A}||_1$. Using this formulation they extract the semantic similarities by the Fisher criterion. Fisher, i.e. $\Omega(\mathbf{A})$, aims to maximize the discrimination of each group of data, and simultaneously minimize the inter-group discrimination. MLDL is an extended version that extracts both the visual and semantic similarities in sparse space. This methos utilizes $||\mathbf{P}^{\top}\mathbf{X}-\mathbf{D} \mathbf{A}||_F^2+||\mathbf{Q}-\mathbf{W}\mathbf{A}||_F^2+||\mathbf{A}||_1$ formula which is somehow similar to our approach with some differences. The second term, $||\mathbf{Q}-\mathbf{W}\mathbf{A}||_F^2$, is where varies from our method. Here, the algorithm represents the $\mathbf{Q}$ matrix containing the semantic information. In fact, $\mathbf{Q}\in \mathbb{R}^{N \times K}$ is a binary matrix ($N$ the number of train samples, and $K$ the number of prototypes). This matrix measures the semantical correspondence of any prototype to the training data. The drawback of $\mathbf{Q}$ is that it is a binary relation, so cannot represent the similarity rate of the data. It assigns 1 if the prototype and the training sample share the same label set, while it could be partially true for many couples. In our method, we learn the semantic similarity of any prototype, while here it is prior knowledge. Moreover, it uses F-norm which is not appropriate for the label loss function and we utilized the hinge loss function instead (we explain the reasons later). Besides, MSFS concentrates on sparse coding for feature extraction by $||\mathbf{Y}-\mathbf{V}\mathbf{B}||_F^2+\Gamma(\mathbf{V})+||\mathbf{X}\mathbf{W}-V||_F^2+\Omega(\mathbf{W})+\lambda(\mathbf{W})$.\: The initial term focuses on dictionary learning for semantic representation, through minimizing the distance of $\mathbf{V}\mathbf{B}$ and $\mathbf{Y}$, where $\mathbf{B}$ is the dictionary and $\mathbf{V}$ is the coefficients matrix. Similar to MLDL it exploits Frobenius-norm for semantic similarity extraction. One more point, in the third term of the objective function it finds a $W$ matrix that its multiplication in $\mathbf{X}$ (training data) reconstructs $\mathbf{V}$. In fact, MSFS through $\mathbf{W}$ estimates the $\mathbf{V}$ coefficients that its multiplication in $\mathbf{B}$ provides the estimated labels.

%%%%%%%%%%%%%%%%%%%%%%%%%%%%%% PROPOSED METHOD %%%%%%%%%%%%%%%%%%%%%%%%%%%%% 
\section{Marginalized Coupled Dictionary Learning}
In this section, the proposed approach (MCDL) is discussed in detail. We present the objective function and learning algorithm of MCDL in Sections  \mbox{\ref{sec_obj}} and  \mbox{\ref{sec_outline}} respectively. Then,  two main steps of the learning algorithm, including marginalized coupled sparse coding  \mbox{\ref{sec_sparsecoding}} and visual and semantic dictionary update \mbox{\ref{sec_dict_update}} are discussed.\\
\label{sec_proposedmethod}
%In this section, the proposed approach (MCDL) is discussed in detail. The objective function and learning algorithm for MCDL are presented in Section \ref{sec_outline} and \ref{sec_obj} respectively. Finally, the proposed objective function is discussed in Section \ref{sec_discussion}.
%%%%%%%% OBJECTIVE FUNCTION %%%%%%%%%
\subsection{Objective Function}
\label{sec_obj}
\begin{figure}
\includegraphics[width=0.4\textwidth]{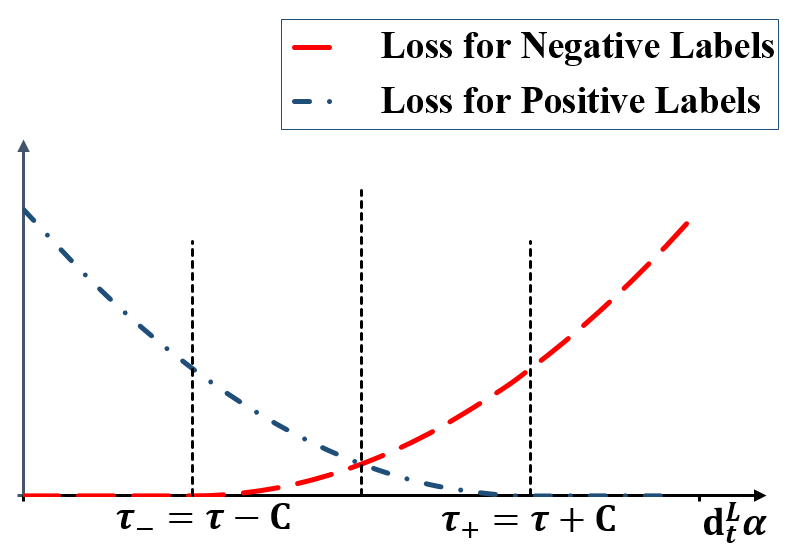} % 
\caption{The squared hinge loss functions for positive and negative labels.}
\label{fig_largemargin}
\end{figure}

In this section, we have presented the objective function of the proposed method (MCDL) in detail.  This method aims to marginalize scores for positive and negative labels. This means that the negative labels with small reconstruction (less than a margin but not zero) do not need to be penalized. Similarly, positive labels whose reconstructions are above a certain margin will not be penalized. Furthermore, to learn sparse semantic prototypes associated with the visual prototypes, MCDL imposes ${\ell}_1$ regularization on the semantic dictionary. Considering these objectives, the empirical cost function for MCDL has been suggested as below:
\begin{equation}
\label{eqn_obj}
\begin{split}
\begin{gathered}
\underset{{\mathbf{D}}^I,{\mathbf{D}}^L,\mathbf{A}}{\mathrm{minimize}}  \: \sum_{i=1}^N  { \left(  \frac{{\mathcal{N}}_{+}^i}{\lambda} {{\parallel x^i-{\mathbf{D}}^I{\alpha}^i \parallel}_2^2} \: + \: {\sum_{t=1}^T{\ell (y_t^i,{\mathbf{d}}_t^L{\alpha}^i)}}\right)} \: \\ + \: \sum_{t=1}^T{{ \beta}_1 {\parallel {\mathbf{d}}_t^L  \parallel}_1 } \qquad \: s.t \: {\parallel {\alpha}^i \parallel}_1 \leq {\beta}_0, \: {\alpha}_k^i \geq 0,  \: \\ {{ \parallel  {d}_{k}^{I}  \parallel }_{2}} \leq 1, \: 0 \leq {{\mathbf{d}}_{t,k}^L} \leq v,  \: \forall i,t,k,
\end{gathered}
\end{split}
\end{equation}
where the first term is the reconstruction term for visual vectors, the second term is the hinge loss function defined over label vectors and the last term is ${\ell}_1$ regularization for the semantic dictionary.  Note that $x^i, \: i \in {1,...,N}$ is the $i^{th}$ training vector and $y_t^i \in \{0,1\}$ is the $t^{th}$ label of it. In the objective function of (\ref{eqn_obj}), $\alpha^i, \: i \in {1,...,N}$ is the shared representation (coupled) over visual and semantic vectors, and ${ \beta}_0 = 1$ is ${\ell}_1$ upper-bound for this representation. We suppose that each visual feature vector is $\ell_2$-normalized, and ${ \beta}_0 = 1$. This will decrease the number of tuning parameters and will increase the generalization of such representation. Moreover, the number of positive labels for each sample (${\mathcal{N}}_{+ }^i$) is employed as a regularizer between visual and semantic terms for each training sample, as well as the tuning parameter of $\lambda$. \\
In the optimization problem of (\ref{eqn_obj}), $v$ is the upper bound for semantic dictionary elements to avoid noisy prototypes (discussed in Section \ref{sec_discussion}). The estimation of $y_t^i$, denoted by ${\mathbf{d}}_t^L{\alpha}^i$, is achieved by multiplying sparse representation into the $t^{th}$ row of the semantic dictionary. The positiveness constraint is also imposed on the coefficients to increase the generalization of learned prototypes. Additionally, the ${\ell}_1$ regularization in the last term aims to learn the most sparse semantic dictionary with annotation power (the second term). Furthermore, ${\ell(.,.)}:\mathbb{R}\times \mathbb{R}\rightarrow\mathbb{R}$ is the squared hinge loss function which is defined as below:
\begin{equation}
\label{eqn_hingeloss}
\ell(y_t^i,{\mathbf{d}}_t^L{\alpha}^i) =  {[\max {(0, \: \mathbf{C} -(2 y_t^i - 1) ( {\mathbf{d}}_t^L{\alpha}^i-{\tau} ))}]}^2, 
\end{equation}
where $ \tau$ is a constant threshold value and $2 \mathbf{C}$ is the desired gap between score values for positive and negative samples of a label. Figure \ref{fig_largemargin} shows the loss function of Equation (\ref{eqn_hingeloss}). As can be seen, the loss will be zero for a label if computed score satisfies the margin values. For violated labels, the loss will be computed using squared loss based on its distance to the equivalent margin.
%%%%%%%% OUTLINE %%%%%%%%%
\subsection{Learning Algorithm}
\label{sec_outline}
Algorithm \ref{algo_learning} presents the procedure of the suggested optimization technique to solve the optimization problem of ~(\ref{eqn_obj}). This algorithm consists of three main stages:\\
1) \textbf{Data Normalization}: in this stage, feature vectors are normalized to have unit $\ell_2-norm$. This is useful to trade-off between the visual and semantic modalities (the first and second terms of Equation (\mbox{\ref{eqn_obj}})) for each sample using a regularizer ($\lambda$) and its labels counts (${\mathcal{N}}_{+ }^i$). Such normalization can also result in more consistent sparse coding in the train and test stage.  \\
2) \textbf{Initialization}: this stage has two steps. In step \hyperref[itm:step21]{2.1}, visual prototypes are initialized by solving the first term of (\ref{eqn_obj}), in which weights are ignored. In the first stage of Step \hyperref[itm:step22]{2.2}, the sparse representations of training samples are calculated over the visual dictionary obtained by solving the optimization problem of the Equation of Step \hyperref[itm:step21]{2.1} of Algorithm \ref{algo_learning}. In other words, visual prototypes are ${\ell}_2$-normalized, and the semantic dictionary elements must be lower than or equal to $v$. \\
3) \textbf{Optimization}: The problem of Equation (\ref{eqn_obj}) is not jointly convex w.r.t dictionaries (${\mathbf{D}}^I$ and ${\mathbf{D}}^L$) and sparse coefficients $\mathbf{A}=[{\alpha}^1, ..., {\alpha}^N]$. However, it is convex w.r.t each of these parameters set when the other one is fixed. Thus, this optimization problem is decomposed into two convex problems including sparse coding and dictionary update. These two steps are applied alternatively to solve the original problem (Step \hyperref[itm:step3]{3} of Algorithm \ref{algo_learning}). These steps are discussed in the following sections.
\justify
%\IncMargin{0.5em}
\begin{algorithm}
\caption{Marginalized Coupled Dictionary Learning}\label{algo_learning}
\SetKwData{Left}{left}\SetKwData{This}{this}\SetKwData{Up}{up}
\SetKwFunction{Union}{Union}\SetKwFunction{FindCompress}{FindCompress}
\SetKwInOut{Input}{Input}\SetKwInOut{Output}{Output}
\Input{Set of training images $\mathbf{X}=\left[ x^1,\dots,x^N\right]$ and their corresponding label vectors $\mathbf{Y}=\left[y^1,\dots,y^N\right]$.\\ Discriminative regularization $\lambda$.\\
The number of prototypes $K$.\\
The number of repeats $R$. }
\Output{Learned prototypes $\mathbf{D}^I$ and $\mathbf{D}^L$.}
\BlankLine 
All necessary notations have been introduced in Section \ref{sec_notations}.\\
\item[\textbf{1.}]\label{itm:step1}
\textbf{Data Normalization}\\
\begin{itemize}
\item{ $x^i \leftarrow  \frac{x^i}{{||x^i||}_2}, \forall i \in \{1,\dots,n\}.$}
\end{itemize}
\item[\textbf{2.}]\label{itm:step2}
\textbf{Initialization}\\
\begin{enumerate}
\item[\textbf{2.1.}]\label{itm:step21} Visual Dictionary Initialization
	\begin{itemize}
	\item{Initialize $\mathbf{D}^I$ by $K$ using k-means clustering .}
	\item{Solve below optimization problem by alternate updating of $\mathbf{A}$ and $\mathbf{D}^I$:\\
%\begin{equation}
%\label{eqn_upd}
%\begin{gathered}
%\underset{{\mathbf{D}}^I,\mathbf{A}}{\mathrm{minimize}}  \: \sum_{i=1}^N  { {{\parallel x^i-{\mathbf{D}}^I{\alpha}^i \parallel}_2^2}} \\ s.t {\parallel {\alpha}^i \parallel}_1 \leq 1, {\alpha}_k^i \geq 0, {{ \parallel  {d}_{k}^{I}  \parallel }_{2}} \leq 1, \: \forall i,k.
%\end{gathered}
%\end{equation}
$\quad\quad\underset{{\mathbf{D}}^I,\mathbf{A}}{\mathrm{minimize}}  \: \sum_{i=1}^N  { {{\parallel x^i-{\mathbf{D}}^I{\alpha}^i \parallel}_2^2}}$ \\ 
$\quad\quad s.t {\parallel {\alpha}^i \parallel}_1 \leq { \beta}_0 , {\alpha}_k^i \geq 0, {{ \parallel  {d}_{k}^{I}  \parallel }_{2}} \leq 1 , \: \forall i,k.$
	}
	\end{itemize}
\item[\textbf{2.2.}]\label{itm:step22} Semantic Dictionary Initialization
	\begin{itemize}
	\item{Calculate the sparse representations \\ $\mathbf{A}= [{\alpha}^1,...,{\alpha}^N]$ based on the visual dictionary.}
	\item{${{d}_{k}^{L}}  \leftarrow \:  \min (v, \: \frac{\sum_{i=1}^N  {{\alpha}_k^i y^i}}{ \sum_{i=1}^N  {{({\alpha}_k^i)}^2}}), \: \forall k \in \{1,\dots,K\}.$ }
	\end{itemize}
\end{enumerate}
\item[\textbf{3.}]\label{itm:step3}
\textbf{Optimization}\\
Solve Equation ~(\ref{eqn_lars}) by alternate updating of $\mathbf{A}$ and $\mathbf{D}^C$ while the other is assumed to be fixed:\\
\For{$r \leftarrow 1$ \KwTo $R$}{
\begin{enumerate}
\item[\textbf{3.1.}]\label{itm:step31}
Update representations $\mathbf{A}=[{\alpha}^1,...,{\alpha}^N]$ using \\ proposed MCSC technique ($\forall i \in \{1,\dots,N\}$):\\
	\For{$s \leftarrow 1$ \KwTo $S$}{
	\begin{itemize}
	\item{Calculate ${\tilde{y}}_t^{i}[s-1]$ as an estimate for \\ label scores using ~(\ref{eqn_y_hat}).}
	\item{Update ${\hat{\alpha}}_t^{i}[s]$ by solving ~(\ref{eqn_lars}). }
	\end{itemize}
	}
	${\alpha}^i \leftarrow  {\hat{\alpha}}_t^{i}[S].$
\item[\textbf{3.2.}] \label{itm:step32}
Update each column of ${\mathbf{D}}^{I}$ and ${\mathbf{D}}^{L}$ using \\ Algorithm (\ref{algo_du}). 
\end{enumerate}
}
\end{algorithm}
%\DecMargin{0.5em}
%%%%%%%% MARGINALIZED COUPLED SPARSE CODING %%%%%%%%%
\subsection{Marginalized Coupled Sparse Coding}
\label{sec_sparsecoding}
In this section, proposed Marginalized Coupled Sparse Coding (MCSC) (Step \hyperref[itm:step31]{3.1} of Algorithm \ref{algo_learning}) is presented to solve the sparse coding problem of Equation ~(\ref{eqn_obj}) when the dictionaries are fixed (Section \ref{sec_sparsecoding}). Indeed, MCSC is a fast method based on LARS (Least Angle Regression) technique (also known as LARS-Lasso) \cite{efron2004least, mairal2010online}. \\
When the dictionaries (${\mathbf{D}}^I$ and ${\mathbf{D}}^L$) in the optimization problem of Equation (\ref{eqn_obj}) are fixed, the problem can be rewritten w.r.t each training sample (${\alpha}^i$) individually as below: 
\begin{equation}
\label{eqn_sparsecoding}
\begin{gathered}
\underset{{\alpha}^i}{\mathrm{minimize}}  \:  f({\alpha}^i) = {{\parallel x^i - {\mathbf{D}}^I{\alpha}^i \parallel}_2^2} \: + \: \frac{\lambda} {{\mathcal{N}}_{+}^i} \sum_{t=1}^T{{{\xi}_t^{i}}^2} \\
s.t \: {\parallel {\alpha}^i \parallel}_1 \leq { \beta}_0, \: {\alpha}_k^i \geq 0, \\  
{\xi}_t^{i} \geq  \mathbf{C} - (2 y_t^i - 1) ( {\mathbf{d}}_t^L{\alpha}^i-{\tau}), \: {\xi}_t^{i} \geq 0,
 \end{gathered}
\end{equation}
where ${\xi}_t^{i}$ is the slack variable which measures margin violation for $t^{th}$ label of the $i^{th}$ training data. According to the constraints in Equation (\ref{eqn_sparsecoding}), we have:
\begin{equation}
\label{eqn_xi_t}
\begin{split}
\begin{gathered}
{\xi}_t^{i} = \max {(0, \: \mathbf{C} -(2 y_t^i - 1) ( {\mathbf{d}}_t^L{\alpha}^i-{\tau} ))}.
\end{gathered}
 \end{split}
\end{equation}
The problem of Equation (\ref{eqn_sparsecoding}), which is equivalent to the first and second terms of Equation (\ref{eqn_obj}), is a constrained quadratic optimization problem that can be solved using quadratic programming techniques. It is noticeable that this problem must be solved for all training samples in each iteration of the whole optimization that is time-consuming. This motivates us to investigate a simpler and faster iterative coupled sparse coding based on the LARS, called MCSC, to obtain an approximate solution for the problem of Equation (\ref{eqn_sparsecoding}). Lasso is an effective and fast method to solve traditional sparse coding problems. To describe MCSC, suppose that ${\hat{\alpha}}^{i}[s-1]$ is the sparse coefficient vector obtained at the previous iteration of MCSC, one can provide a new approximate (${\hat{\alpha}}^{i}[s]$) by solving (see Lemma 1 in \ref{sec_lemma1}):
\begin{equation}
\label{eqn_lars_sc}
\begin{gathered}
{\hat{\alpha}}^{i}[s] \triangleq \underset{{\alpha}^{i}}{\mathrm{arg min}}  \:  g({\alpha}^{i}) = {{\parallel x^i - {\mathbf{D}}^I{{\alpha}^{i}} \parallel}_2^2} \: \\ + \: \frac{\lambda}{{\mathcal{N}}_{+}^i} \sum_{t=1}^T {{({\tilde{y}}_t^{i}[s-1]-{\mathbf{d}}_t^L {\alpha}^{i})}^2}
\\  \: s.t \: {\parallel \alpha \parallel}_1 \leq { \beta}_0, \: {\alpha}_k^i \geq 0,
\end{gathered}
\end{equation}
where ${\tilde{y}}_t^{i}[s-1]$ ($t \in \{1\dots,T\}$) is defined based on current approximate of sparse coefficients (${\hat{\alpha}}^{i}[s-1]$) and its corresponding label penalties (${\hat{\xi}}_t^{i}[s-1], \forall t \in \{1\dots,T\}$) as below:
\begin{equation}
\label{eqn_y_hat}
\begin{split}
\begin{gathered}
{\tilde{y}}_t^{i}[s-1] = \left \{ \begin{array}{rcl} {\mathbf{d}}_t^L {\hat{\alpha}}^{i}[s-1], & \mbox{if}  &  {\hat{\xi}}_t^{i}[s-1]=0,  \\
{\tau}+(2 y_t^i - 1) \mathbf{C} & \mbox{if}  &  {\hat{\xi}}_t^{i}[s-1]>0.
\end{array} \right .
\end{gathered}
 \end{split}
\end{equation}
The optimization problem of (\ref{eqn_lars_sc})  can be reformulated as:
\begin{equation}
\label{eqn_lars}
\begin{gathered}
\underset{{\alpha}^{i}}{\mathrm{arg min}}  \: {{ \left \|  \left[ \begin{array}{c} {x^i} \\ \sqrt{\frac{\lambda}{{\mathcal{N}}_{+}^i}} {{\tilde{y}}_t^{i}[s-1]}\end{array}\right]- \left[ \begin{array}{c}{\mathbf{D}}^I \\ \sqrt{\frac{\lambda}{{\mathcal{N}}_{+}^i}} {\mathbf{D}}^L\end{array}\right]  {{\alpha}^{i}} \right \| }_2^2} \\
\: s.t \: {\parallel {\alpha}^i \parallel}_1 \leq 1, \: {\alpha}_k^i\geq 0,
\end{gathered}
\end{equation}
which is equivalent to a sparse coding problem with positiveness constraint on the coefficients vector. This optimization problem can be solved effectively using LARS. Step \hyperref[itm:step31]{3.1} of Algorithm \ref{algo_learning} summarizes MCSC algorithm presented to solve Equation (\ref{eqn_sparsecoding}). In each iteration of MCSC, the optimization problem of (\ref{eqn_lars}) is solved using LARS based on the current estimate for sparse coefficients provided at the previous iteration. In this paper, we have repeated this step four times ($S=4$). Initial approximate is obtained by supposing that ${\hat{\xi}}_t^{i}[0]>0$ for all labels in Equation (\ref{eqn_y_hat}). We have presented Lemma 1 in \ref{sec_lemma1} to prove the convergence of the MCSC algorithm.
%\justify
%\IncMargin{0.5em}
\begin{algorithm}%[H]
\SetKwRepeat{Do}{do}{while}
\SetKwData{Left}{left}\SetKwData{This}{this}\SetKwData{Up}{up}
\SetKwFunction{Union}{Union}\SetKwFunction{FindCompress}{FindCompress}
\SetKwInOut{Input}{Input}\SetKwInOut{Output}{Output}
\Input{Set of normalized training images ${\mathbf{X}}=\left[ x^1,\dots,x^N\right]$ and their corresponding label vectors ${\mathbf{Y}}=\left[y^1,\dots,y^N\right]$.\\
Current learned prototypes ${\mathbf{D}}^I$ and ${\mathbf{D}}^L$.\\ 
Regularization parameter ${\beta}_1$ for $l_1-norm$.\\
Sparse representations $\mathbf{A}=[{\alpha}^1,...,{\alpha}^N]$.}
\Output{Updated ${\mathbf{D}}^I$ and ${\mathbf{D}}^L$.}
\BlankLine
\textbf{Visual and Semantic Dictionary Update}\\
\For{$j \in \{1, \dots, K\}$ at random}
{
\begin{itemize}
\item{Update ${k}^{th}$ Visual Prototype:\\}
\begin{itemize}
	\item{ $z_k^i =  x^i - (D^I{\alpha}^i -  {d_k^I}{\alpha}_k^i), \: \forall i \in \{1,\dots,N\}.$ }
	\item{${{\hat{d}}_{k}^{I}}  \leftarrow \:  \frac{\sum_{i=1}^N  {{\mathcal{N}}_{+}^i {\alpha}_k^i z_k^i}}{ \sum_{i=1}^N  {{\mathcal{N}}_{+}^i {({\alpha}_k^i)}^2} }, \: \forall k \in \{1,\dots,K\}.$ }
\item{${d_{k}^{I}}  \leftarrow \:  \frac{{{\hat{d}}_{k}^{I}}}{{{||{\hat{d}}_k^I||}_2}}.$}
\end{itemize}
\item{Update ${k}^{th}$ Semantic Prototype:\\}
	\For{$t \leftarrow 1$ \KwTo $T$}{
	\begin{itemize}
	\item{$q_k^i =  D_t^L {{\alpha}^i} -  {{\mathbf{d}}_{t, k}^L} {{\alpha}_k^i},\: \forall i \in \{1,\dots,N\}.$}
	\item{Update ${\mathbf{d}}_{t, k}^L$ by solving ~(\ref{eqn_sdu1}) }
	\end{itemize}
	}
\end{itemize}
}
\caption{Dictionary Update}\label{algo_du}
\end{algorithm}
%\DecMargin{0.5em}
%%%%%%%% DICTIONARY UPDATE %%%%%%%%%
\subsection{Dictionary Update}
\label{sec_dict_update}
Consider solving the optimization problem of ~(\ref{eqn_obj}) when the sparse coefficients ($\mathbf{A}=[{\alpha}^1, ..., {\alpha}^N]$) are fixed. In this case, this problem is equivalent to optimize visual and semantic dictionaries separately. We have utilized a randomized coordinate descent algorithm based on warm restart (current parameters) to update prototypes (columns) of both dictionaries in a random sequence, summarized in Algorithm \ref{algo_du}. In this section, we have presented the proposed methods to solve these two disjoint dictionary learning problems to optimize semantic and visual prototypes.
%%% VISUAL DICTIONARY UPDATE %%%
\subsubsection{Visual Dictionary Update}
When keeping the sparse coefficients fixed, the optimization problem of ~(\ref{eqn_obj}) w.r.t visual dictionary is equivalent to solve the below problem:
\begin{equation}
\label{eqn_vdu}
\begin{gathered}
\underset{{\mathbf{D}}^I}{\mathrm{minimize}}  \: \sum_{i=1}^N  {\mathcal{N}}_{+}^i {{\parallel x^i-{\mathbf{D}}^I{\alpha}^i \parallel}_2^2} \quad
\: s.t   \: {{ \parallel  {d}_{k}^{I}  \parallel }_{2}} \leq 1.
\end{gathered}
\end{equation}
This objective function is indeed a weighted form of the traditional dictionary learning problem. One of the most used approaches to solve this problem is the block coordinate descent approach, in which prototypes are optimized individually while keeping the others fixed \cite{mairal2010online, wang2017cross}. Taking the gradient of (\ref{eqn_vdu}) w.r.t $d_{k}^I$ and setting it equal to zero, we have:
\begin{equation}
\label{eqn_vdu1}
\begin{split}
\begin{gathered}
{{\hat{d}}_{k}^{I}}  \leftarrow \:  \frac{\sum_{i=1}^N  {{\mathcal{N}}_{+}^i {\alpha}_k^i z_k^i}}{ \sum_{i=1}^N  {{\mathcal{N}}_{+}^i {({\alpha}_k^i)}^2} },
\end{gathered}
\end{split}
\end{equation}
where $z_k^i =  x^i - (D^I{\alpha}^i -  {d_k^I}{\alpha}_k^i)$ is residual of the $i^{th}$ input vector w.r.t other prototypes, and ${{\hat{d}}_{k}^{I}}$ is the optimum of (\ref{eqn_vdu}) without considering its constraint. This can be shown that solving constrained optimization (\ref{eqn_vdu}) w.r.t to the $k^{th}$ prototype (column) of the visual dictionary (i.e., $d_{k}^I$), when the other prototypes hold fixed, is equivalent to solve unconstrained one, followed by an ${ \ell}_2-norm$ normalization. It is worth mentioning that ${\mathcal{N}}_{+}^i $ (the number of positive labels for $i^{th}$ sample) acts as weights in updating visual prototypes. This means that samples with more labels will have a greater impact on the optimized visual prototypes because they are probably annotated with complete labels. In the marginalized sparse coding problem of (\ref{eqn_lars_sc}), these weights play the role of normalizer to make a balance between visual and semantic loss functions.
%%% SEMANTIC DICTIONARY UPDATE %%%
\subsubsection{Semantic Dictionary Update}
\begin{figure}
   \subfloat[]{\label{rev}
      \includegraphics[width=0.22\textwidth]{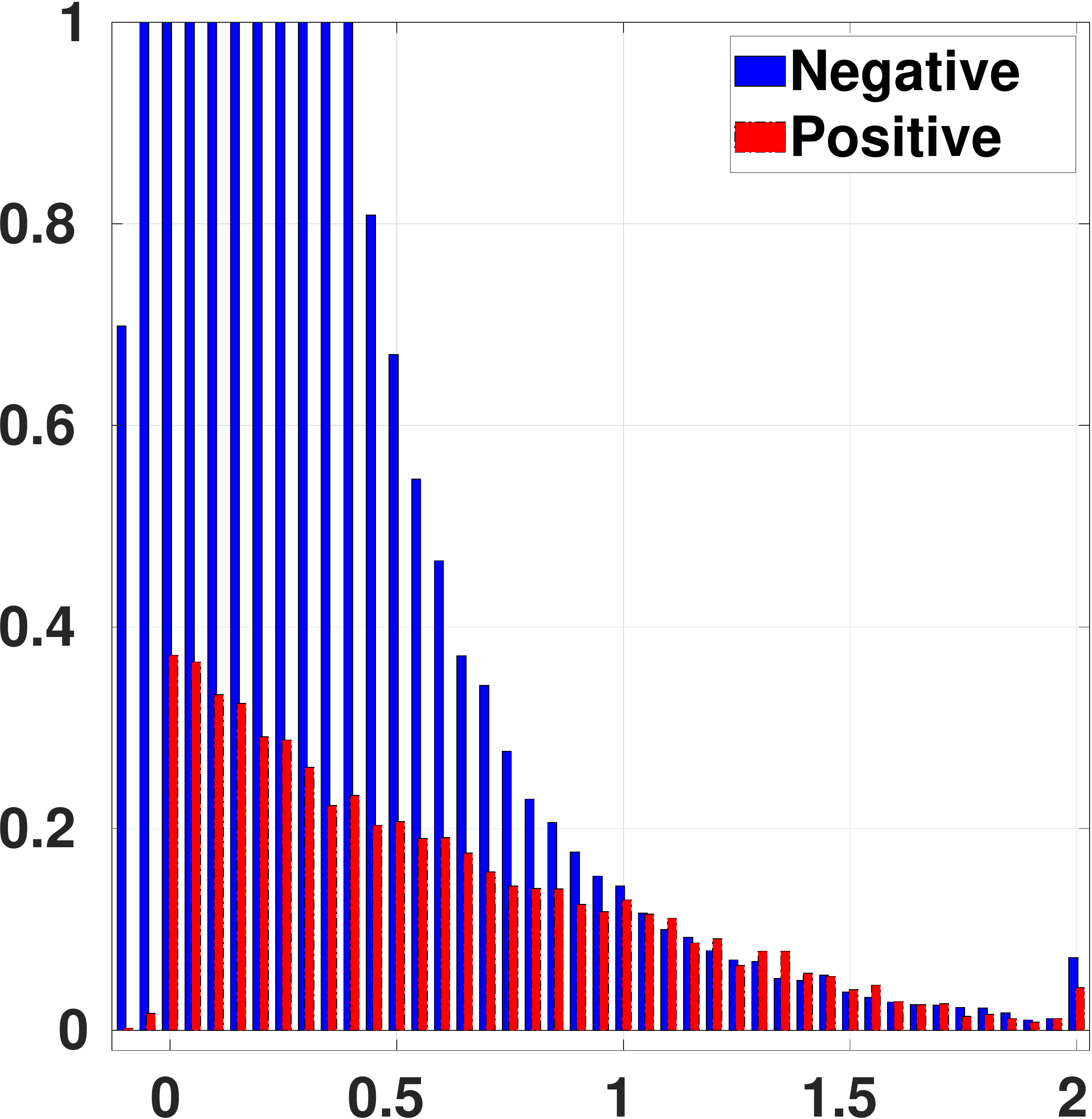}}
      %\label{fig_margin1}
~
   \subfloat[]{\label{rev_sol}
      \includegraphics[width=0.23\textwidth]{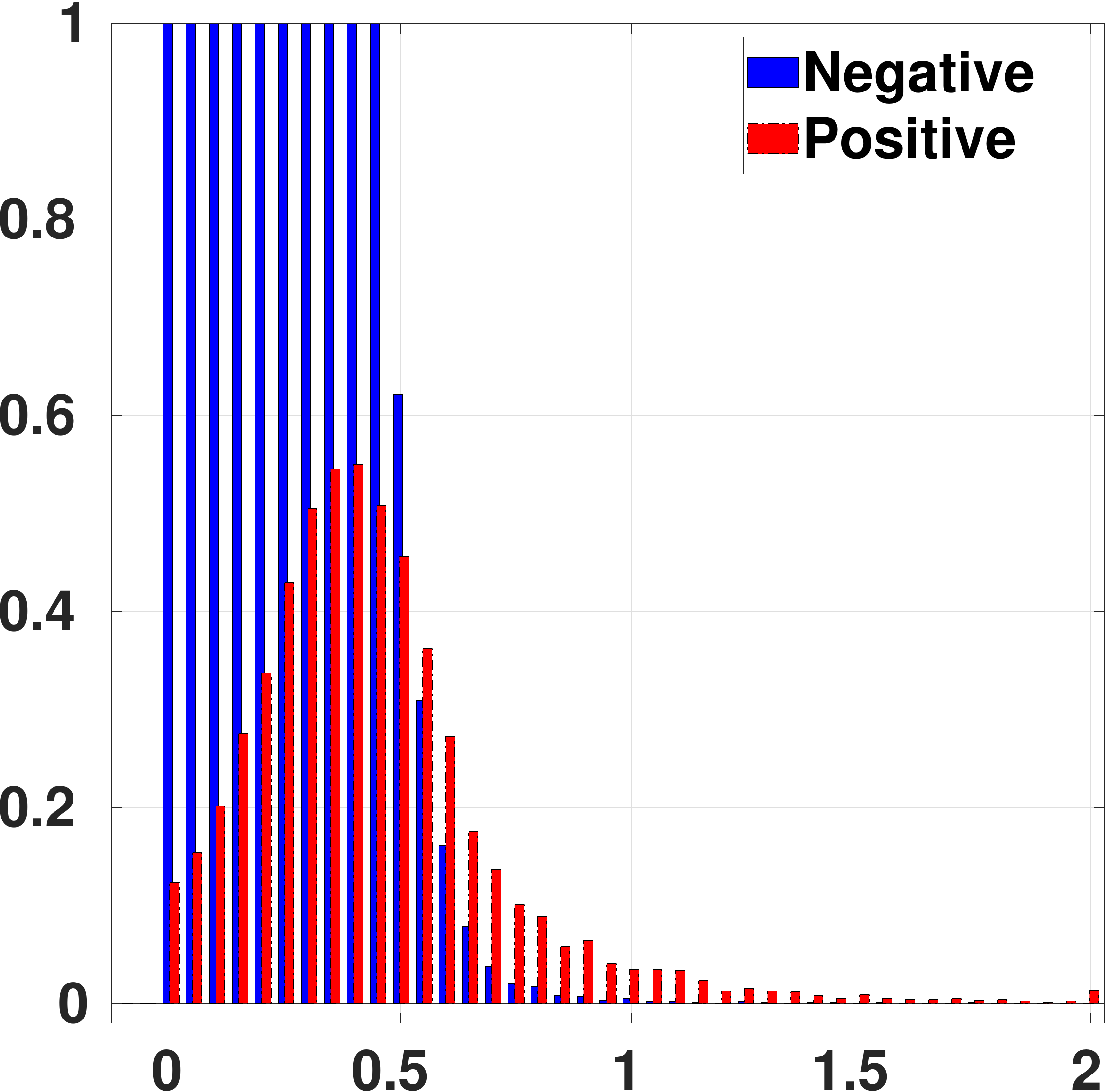}}
      %\label{fig_margin2}
\caption{The distribution of scores (scaled for visualization purpose) for positive and negative labels generated (a) using squared loss function and (b) using MCDL with hinge loss. This result has been provided using IAPRTC-12 dataset and 4000 prototypes.}
\label{fig_margin}
\end{figure}

If the sparse coefficients are given, the optimization problem of ~(\ref{eqn_obj}) w.r.t the semantic dictionary turns into $T$ independent convex problems (one per each label) as below:
\begin{equation}
\label{eqn_sdu}
\begin{gathered}
\underset{{\mathbf{d}}_t^L}{\mathrm{minimize}}  \: \sum_{i=1}^N  {{({\xi}_t^{i})}^2}  + \: {  { \beta}_1 {\parallel {\mathbf{d}}_t^L  \parallel}_1 } \quad
\: s.t  \:\: 0 \leq {{\mathbf{d}}_{t,k}^L} \leq v, \forall k, \\
 {\xi}_t^{i}  \geq  \mathbf{C} - (2 y_t^i - 1) ( {\mathbf{d}}_t^L{\alpha}^i-{\tau}) , \: {\xi}_t^{i} \geq 0, \forall i.
\end{gathered}
\end{equation}
This problem can be seen as a modification of support vector machine with ${ \ell}_1-norm$ regularization, where hinge loss is replaced with squared hinge loss. In the proposed semantic dictionary learning approach, ${ \ell}_1-norm$ regularization can act as a prototype selection for each label, meaning that just a small number of prototypes can be representative for each label. The relation between threshold ($\tau$), margin ($\mathbf{C}$), and semantic elements upper bound ($v$) are discussed in the next section. The problem of (\ref{eqn_sdu}) is a quadratic optimization problem that can be solved using quadratic programming approaches, though it needs high computational time and memory. Since this optimization problem should be solved in each iteration of the dictionary learning for all labels, we have proposed a simple and fast approach based on block coordinate descent. Each of these $T$ convex problems of (\ref{eqn_sdu}) admits separable constraints (${ \ell}_1-norm$) in the updated blocks (${{\mathbf{d}}_{t,k}^L}, \forall k \in \{1,\dots, K\}$). So, the convergence of the proposed coordinate descent based method is guaranteed \cite{wright2015coordinate}. To optimize ${\mathbf{d}}_{t,k}^L$ which is $k^{th}$ element (column) of semantic dictionary for $t^{th}$ label (row) using block coordinate descent when the other variables are fixed, we should solve:
\begin{equation}
\label{eqn_sdu1}
\begin{gathered}
\underset{{\mathbf{\hat{d}}}_{t, k}^L}{\mathrm{minimize}}  \: \sum_{i \in \{i \mid {\alpha}_k^i \neq 0\} }  {{({\xi}_t^{i})}^2}  + \: {{ \beta}_1 |{\mathbf{\hat{d}}}_{t, k}^L}| + \rho \: {{\parallel {\mathbf{\hat{d}}}_{t, k}^L - {{\mathbf{d}}_{t, k}^L} \parallel}_2^2} \\
 \quad \: s.t  \: 0 \leq {{\mathbf{\hat{d}}}_{t, k}^L} \leq v, \:
  \\ {\xi}_t^{i} = \max {(0, \: \mathbf{C} -(2 y_t^i - 1) ({\mathbf{\hat{d}}}_{t, k}^L{{\alpha}_k^i + {\mathbf{q}}_t^L -{\tau}} ))}.
\end{gathered}
\end{equation}
where $q_k^i =  D_t^L {{\alpha}^i} -  {{\mathbf{d}}_{t, k}^L} {{\alpha}_k^i}$  is the score (regression) of $t^{th}$ tag of $i^{th}$ label vector using other semantic prototypes and ${\mathbf{\hat{d}}}_{t, k}^L$ is the new estimate for  ${\mathbf{d}}_{t, k}^L$.  \\
The cost function of ~(\ref{eqn_sdu1}) is a single variable optimization problem, which can be solved effectively even using a parallel linear search for all tags simultaneously. Since the optimization problem of ~(\ref{eqn_sdu}) is a convex optimization problem with separable regularization, convergence of the proposed coordinate descent method is guaranteed. \\
Figure \ref{fig_margin} illustrates the distribution of scores (for test samples) based on the squared loss function versus the marginalized MCDL approach. As Figure \ref{fig_margin}-a shows, scores distribution of the positive labels gets biased to zero when the squared loss function is employed. Figure \ref{fig_margin}-b illustrates the impact of hinge loss on the distribution of scores, where there are less interaction between scores of negative and positive samples.
%%%%%%%% DISCUSSION %%%%%%%%%
\subsection{Discussion and Parameters Setting}
\label{sec_discussion_and_parameter_setting}
In the previous sections, we suggested a joint optimization problem to learn visual and semantic dictionaries in a coupled manner. The objective function of (\mbox{\ref{eqn_sdu}}) can be regarded as a modification of ${ \ell}_1-norm$ Support Vector Machine \mbox{\cite{zhu20041, hastie2015statistical}} which was first presented for feature selection in high dimensional feature spaces. In fact, each row of the semantic dictionary is indeed a regression vector over sparse coefficients to predict the associated labels (refer to \mbox{\cite{jiang2011learning}}). The number of prototypes is usually greater than the number of positive samples for each label, which can raise the issue of overfitting. MCDL utilizes ${ \ell}_1$ regularization to learn semantic prototypes as sparse as possible. Another inspiration to use ${ \ell}_1$ regularization is that label vectors are sparse by nature, and each visual prototype could correspond to a few labels. Moreover, we are interested in non-negative semantic prototypes and sparse coefficients for the same reasons. This constraint has been extensively applied in non-negative matrix factorization techniques \mbox{\cite{mairal2010online, rad2018multi}} for a wide range of applications and can produce more localized features.   \\
As mentioned in Section \mbox{\ref{sec_obj}}, \mbox{${\hat{y}}_t^i = $} \mbox{${{\mathbf{d}}_{t}^L}{\alpha}^i$} \mbox{$ \in \mathbb{R}$} (Figure \mbox{\ref{fig_largemargin}}) is the score value of $t^{th}$ label for the $i^{th}$ training data. Due to the constraints of (\mbox{\ref{eqn_obj}}) on sparse coefficients (${\parallel \alpha \parallel}_1 \leq 1$) and semantic dictionary elements ($\: 0 \leq {{\mathbf{d}}_{t,k}^L} \leq v, \forall t,k$), the upper bound value for scores is $\: 0 \leq {\hat{y}}_t^i \leq v, \forall i,t$. Suppose that $v=1$ and $\tau=\mathbf{C}=0.5$. In this case, to obtain zero hinge loss for a positive label, all used prototypes in sparse representation should be exactly $1$ for this label, which is impossible in practice. In other words, chosen values for threshold and margin values impact on appropriate value of $v$. On the contrary, greater values of $v$ can lead to noisy prototypes which can be controlled using tuning parameter of ${ \beta}_1$ to some extent. Therefore, in this paper, we have set $\mathbf{C} \in \{0.25, 0.5\} $,  where $\tau = 0.25 + \frac{\mathbf{C}}{2}$, and $v=5$. \\

Another point is that upper bound for squared hinge loss in the first term of Equation (\ref{eqn_sdu}) is $N_t^{+}{(\tau + \mathbf{C})}^2$ (when ${\mathbf{d}}_{t}^L=\vec{0}$). The second term of this equation guarantees that increasing the value of a semantic dictionary element will be equivalent to decreasing hinge loss at a meaningful level. So, the ${\ell}_1-norm$ of semantic dictionary rows remains correlated with the number of positive samples for the associated label. \\
Finally, for tuning of discriminative and ${\ell}_1$ regularizations, we have selected: $\lambda =  \frac{1}{T} {\eta}^2 $, where $ \eta \in \: \{0.1,\allowbreak 0.25, \allowbreak 0.5, \allowbreak 1, \allowbreak 2, \allowbreak 5, \allowbreak 10\} $) and ${ \beta}_1 \in \{0.05,\:  0.1,\allowbreak 0.15, \allowbreak 0.2, \allowbreak  0.25,  \allowbreak  0.3,  \allowbreak  0.4, \allowbreak 0.5,   \allowbreak  0.75, \allowbreak 1\}$ respectively. Best parameters are chosen by validation technique for each dictionary size. The maximum number of iterations for  Step \hyperref[itm:step3]{3} of Algorithm \ref{algo_learning} is 15. To speed up, just two iterations are applied in the validation step.
%%%%%%%%%%%%%%%%%%%%%%%%%%%%%% ANNOTATION STRATEGY %%%%%%%%%%%%%%%%%%%%%%%%%%%%%
\section{Annotation Strategy}
\label{sec_annotation}
\begin{figure}
  \includegraphics[width=0.49\textwidth]{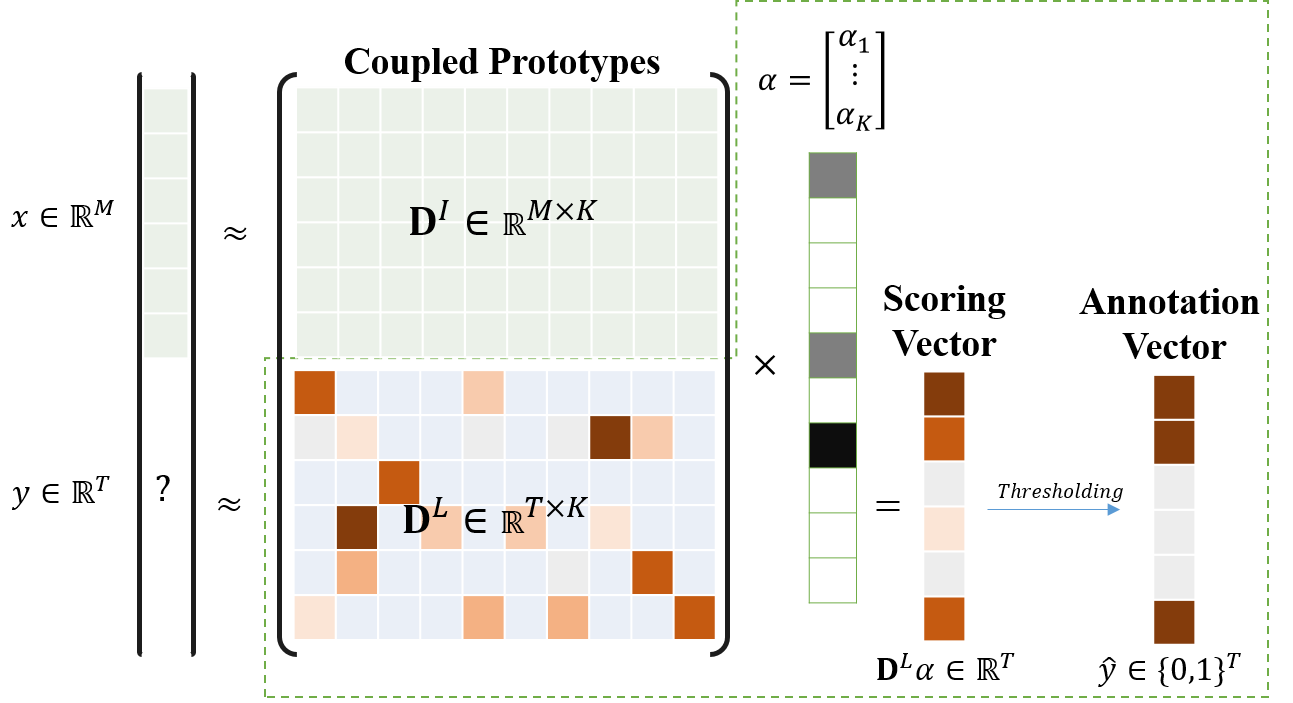}
\caption{The annotation strategy for a query image. The image is annotated with labels that their reconstructions (scores) over semantic dictionary are greater than a given threshold.}
\label{fig_annotationStrategy}
\end{figure}

Figure \ref{fig_annotationStrategy} shows the annotation strategy of a query example. In the first step, the shared sparse representation (which is denoted by $\alpha \in \mathbb{R}^K$) is obtained for the query image, which means $\hat{x} = {\mathbf{D}}^I {\alpha}, \: s.t \: {||{\alpha}||}_1 \leq 1$,  where $x$ is the normalized visual modality. Then, the scoring vector, which shows the similarity of a given image to different labels, is computed as ${\mathbf{D}}^L\alpha\in \mathbb{R}^T$. Finally, we will have $\hat{y}_t = {sign({\mathbf{d}}_t^L{\alpha}-{\tau}_{optimal} )}$, where $\hat{y}_t $ is the prediction for $t^{th}$ label and ${\tau}_{optimal}$ is the optimal threshold for labels prediction. This threshold is computed based on the best $F_1$ measure on training samples for each dataset.

\section{Experimental Results}
\label{sec_expRes}
%%%%%%%%%% FEATURES %%%%%%%%%%%
\hl {\subsection{Datasets and Features}}
\label{sec_Features}
\textbf{Datasets.}
To assess the performance of the proposed method two popular image annotation datasets, IAPRTC-12 \cite{grubinger2006iapr} and ESP-GAME \cite{von2004labeling} are used. Moreover, we utilize another dataset containing one million images from Flickr platform\footnote{https://www.flickr.com}, which could be considered as a more challenging and qualified real-world dataset for this task. We extracted two different subsets from Flickr, called FLICKR-60K and FLICKR-125K, using the following procedure. Similar to \cite{huiskes2010new}, FLICKR-60K was obtained by listing 295 tags which occurred in at least 500 images in the first 100,000 images of the original dataset. We have also removed images with less than two tags, resulting in a dataset with 59083 images. FLICKR-125K was obtained from the first 200,000 images in a way similar to FLICKR-60K. We then split them into train and test sets with a 70-30 ratio. General statistical information of all datasets has been presented in Table \ref{tab_datasets}. \\

\begin{table}
\caption{The number of images (total, train, and test sets) and tags for the datasets.}
\label{tab_datasets}
\begin{tabular}{lccccc}
\hline\noalign{\smallskip}
\textbf{Datasets} & \textbf{Image} & \textbf{Train} & \textbf{Test} & \textbf{Tags}  \\
\noalign{\smallskip}\hline\noalign{\smallskip}
\textbf{IAPRTC-12} & 19627 & 17665 & 1962 & 291 \\ %\citep{grubinger2006iapr} 
\textbf{ESP-GAME} & 20770 & 18689 & 2081 & 268\\ %\citep{von2004labeling} 
\textbf{FLICKR-60K} & 59083 & 41359 & 17724 & 295 \\ %\citep{huiskes2010new}
\textbf{FLICKR-125K} & 124840 & 87388 & 37452 & 568 \\ %\citep{huiskes2010new}
\noalign{\smallskip}\hline
\end{tabular}
\end{table}

%\begin{table}
%\caption{The number of images (total, train, and test sets) and tags for the datasets.}
%\label{tab_datasets}
%\begin{tabular}{lccccc}
%\hline\noalign{\smallskip}
%\textbf{Datasets} & \textbf{Image} & \textbf{Train} & \textbf{Test} & \textbf{Tags} & Stat  \\
%\textbf{IAPRTC-12} & 19627 & 17665 & 1962 & 291 & 5.7, 5, 23 \\ %\citep{grubinger2006iapr} 
%\textbf{ESP-GAME} & 20770 & 18689 & 2081 & 268 & 4.7, 5, 15 \\ %\citep{von2004labeling}
%\textbf{FLICKR-60K} & 59083 & 41359 & 17724 & 295 & 5.27, 4, 40 \\ %\citep{huiskes2010new}
%\textbf{FLICKR-125K} & 124840 & 87388 & 37452 & 568 & 5.88, 5, 64 \\ %\citep{huiskes2010new}
%\end{tabular}
%\end{table}

%\begin{table}
%\caption{Information about the structure of the deep networks.}
%\label{tab_deeps}
%\begin{tabular}{lccc}
%\hline\noalign{\smallskip}
%\textbf{Name} & \textbf{Layers} &\textbf{Dimension} &\textbf{Memory}  \\
%\textbf{VggNet-16} \cite{simonyan2014very}  & 16 & 4048 & 58 M \\ %
%\textbf{ResNet-101} \cite{he2016deep}  & 101 & 2048 & 155 M \\ %
%\textbf{DenseNet-161} \cite{huang2017densely} & 161 & 2208 & 152 M \\ %
%\end{tabular}
%\end{table}
%%%%%%%%%% DATASETS %%%%%%%%%%%

\begin{figure}
\begin{tabular}{cc}
  \makecell{\includegraphics[width=0.44\linewidth]{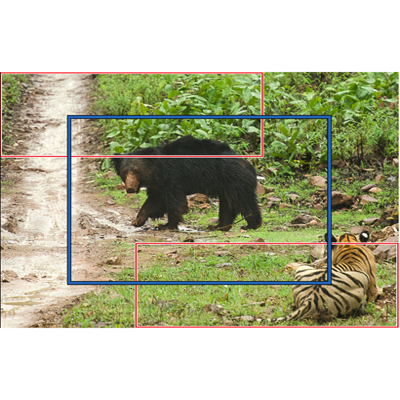}} &  \makecell{\includegraphics[width=0.46\linewidth]{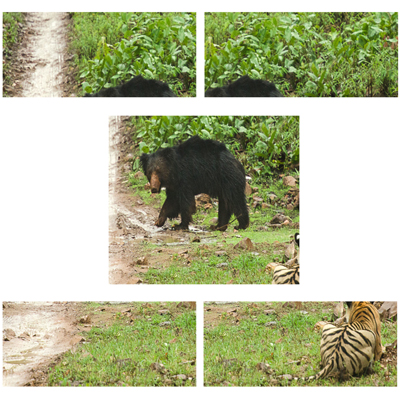}}  \\
\end{tabular}
\caption{The scheme employed to divide an input image before feature extraction using deep models. Five extracted regions have been shown.}
\label{fig_feature}
\end{figure}

\textbf{Features.}
We have employed CNN models that are trained through ImageNet  dataset for object recognition, including VggNet \mbox{\cite{simonyan2014very}}, ResNet \mbox{\cite{he2016deep}}, DenseNet \mbox{\cite{huang2017densely}}, and EfficientNet \mbox{\cite{tan2019efficientnet}} which result in the feature vectors by the dimensionality of 4048, 2048, 2208, and 2560 respectively. To extract these feature vectors, the output of the layer before the last layer is utilized. \\
Random or regular cropping is a common scheme for data augmentation in both training and testing stages \mbox{\cite{krizhevsky2012imagenet, he2016deep, huang2017densely}}. In the training stage, random cropping is widely used to maintain desired image size depending on the network configuration \mbox{\cite{krizhevsky2012imagenet, he2016deep, huang2017densely}}. As a result, cropping techniques such as regular \mbox{\cite{krizhevsky2012imagenet}} or multi-scale cropping \mbox{\cite{he2016deep}} are used in the testing stage to consider all spatial information and prevent downsampling or center cropping to provide network input size. Experiments provided in \mbox{\cite{huang2017densely}} show that the 10-crop strategy (first presented by Krizhevsky et al. \mbox{\cite{krizhevsky2012imagenet}}) outperforms single-crop at test time. \\
 In the standard 10-crop technique, five patches of the same size (the four corners and the center) are extracted as well as their horizontal flips, which results in ten crops. Finally, the predictions of the network are averaged over ten crops in the test stage. We follow the same strategy and obtain five crops (flipping is ignored) for feature extraction of each training and test image. To segment an input image, we crop the central part of the image included in $\frac{2}{3}$ of the whole area, as well as four crops from the corners with the width of $ \frac{2}{3}$ and the height of $ \frac{1}{3}$ (see Figure \mbox{\ref{fig_feature}}). Each segment is then passed to the network and all five extracted feature vectors are averaged to make the final feature vector. Finally, similar to \mbox{\cite{wu2017diverse}}, we apply PCA to reduce the dimensionality of the feature vectors to 200.

%%%%%%%%%% ANALYSIS OF THE DICTIONARY SIZE %%%%%%%%%%%
\subsection{Analysis of the Dictionary Size}

\begin{figure}
\begin{tabular}{lcc}
  \textbf{} &  \textbf{IAPRTC-12} &  \textbf{ESP-GAME}  \\
  \textbf{\rotatebox[origin=c]{90}{DenseNet-161}} & \makecell{\includegraphics[width=.42\linewidth,height=0.30\linewidth]{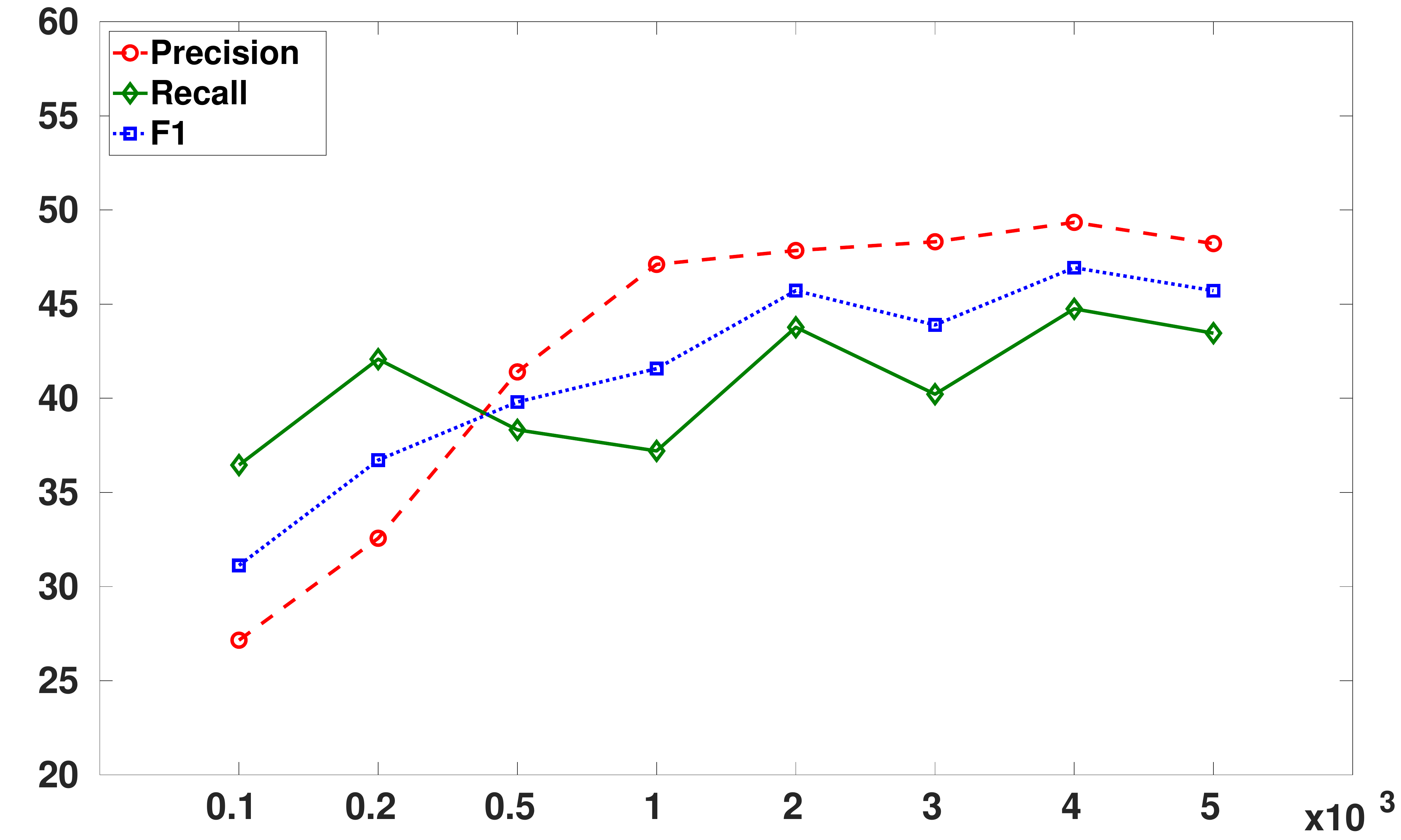}} &  \makecell{\includegraphics[width=.42\linewidth,height=0.30\linewidth]{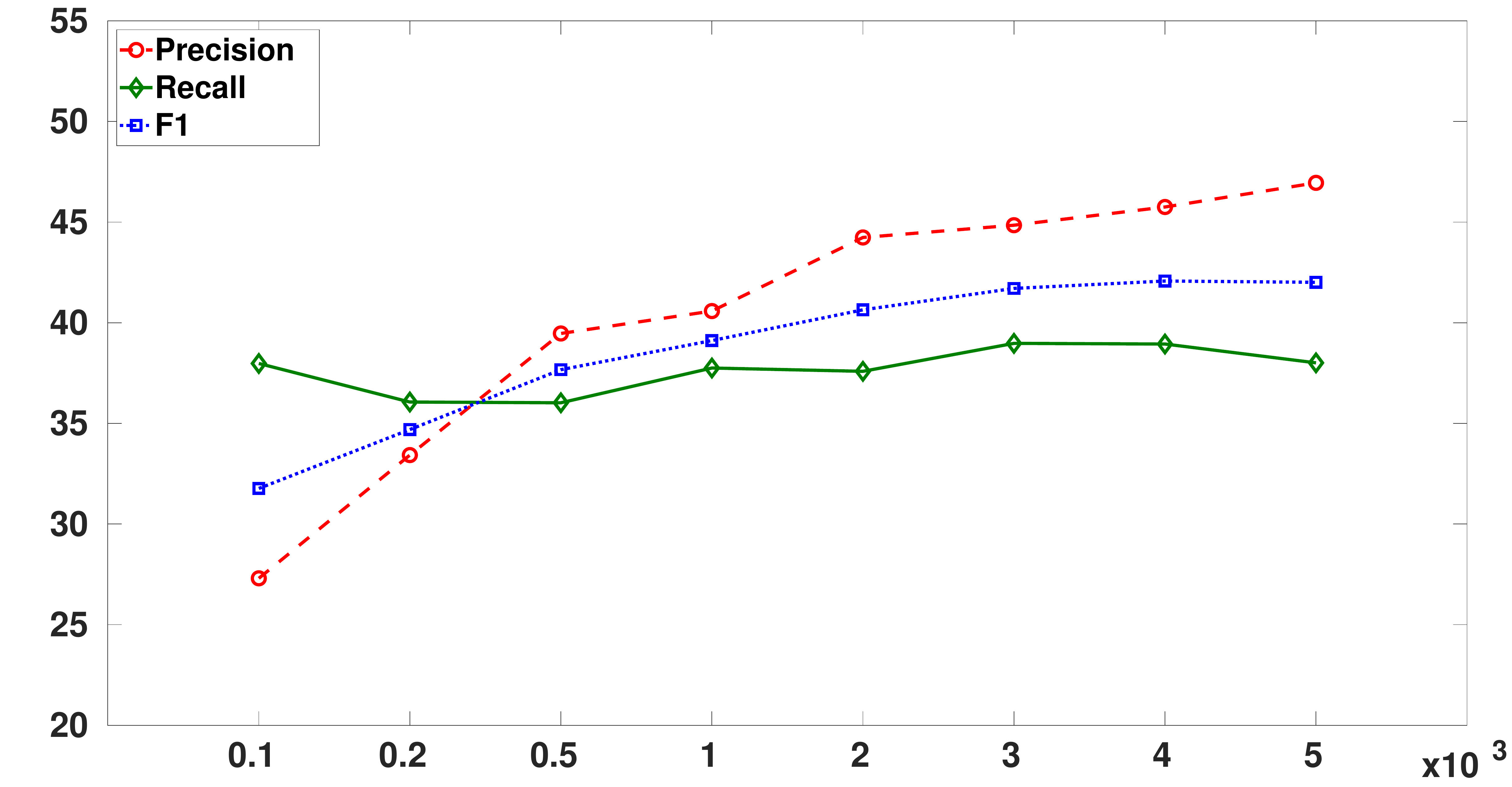}}  \\
  \textbf{\rotatebox[origin=c]{90}{ResNet-101}} &  \makecell{\includegraphics[width=.42\linewidth,height=0.30\linewidth]{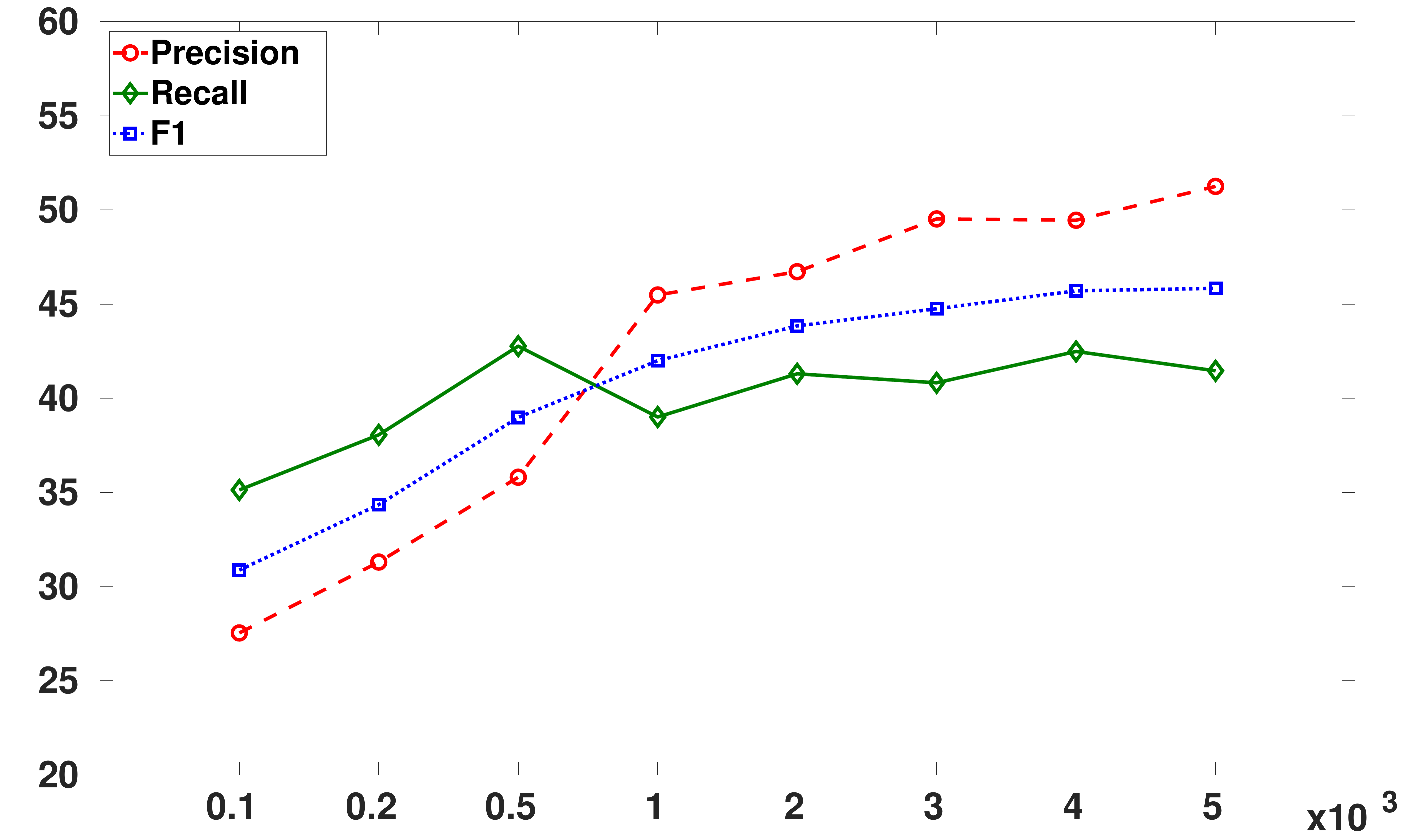}} &  \makecell{\includegraphics[width=.42\linewidth,height=0.30\linewidth]{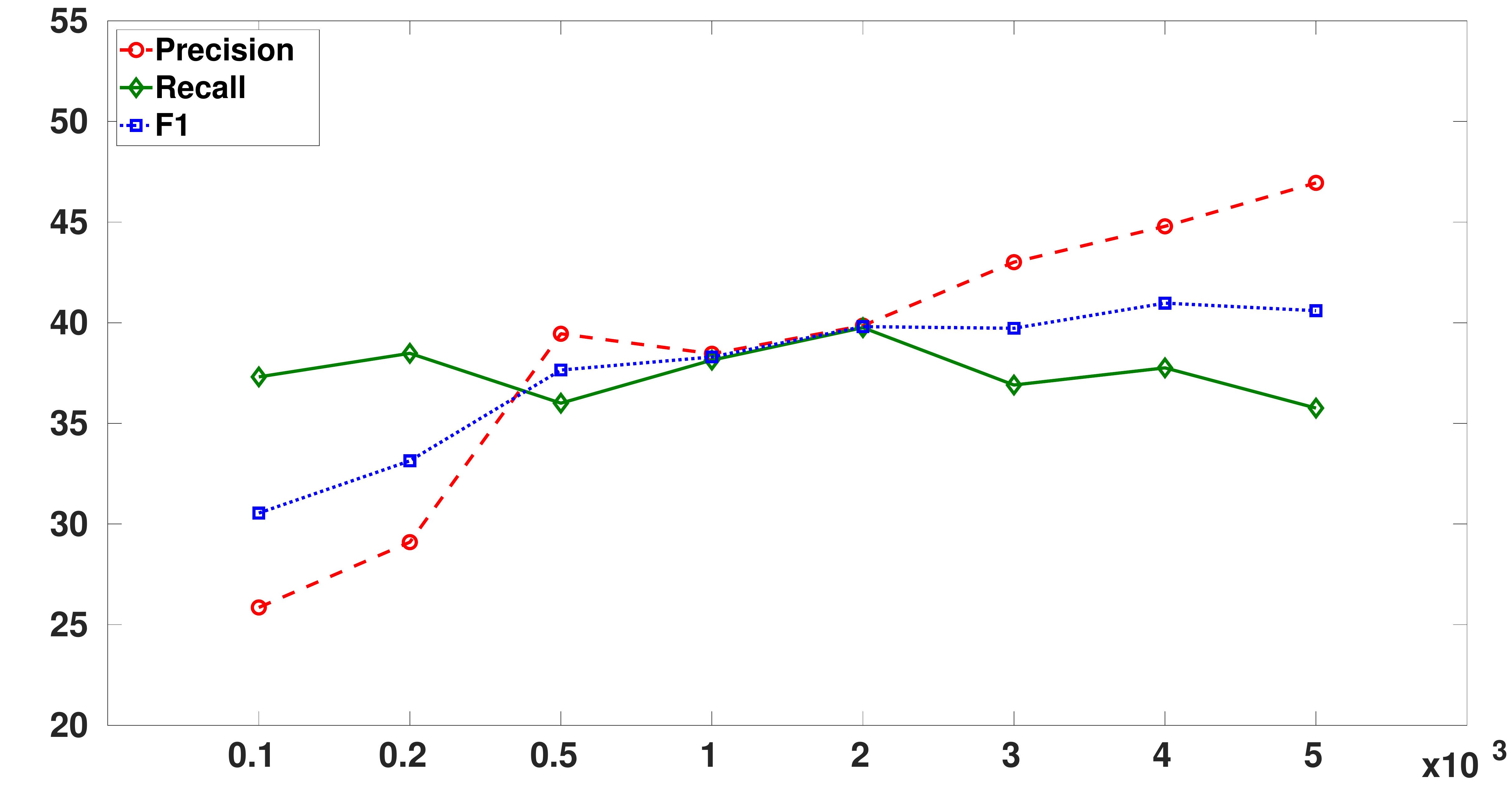}}  \\
  \textbf{\rotatebox[origin=c]{90}{VggNet-16}} &  \makecell{\includegraphics[width=.42\linewidth,height=0.30\linewidth]{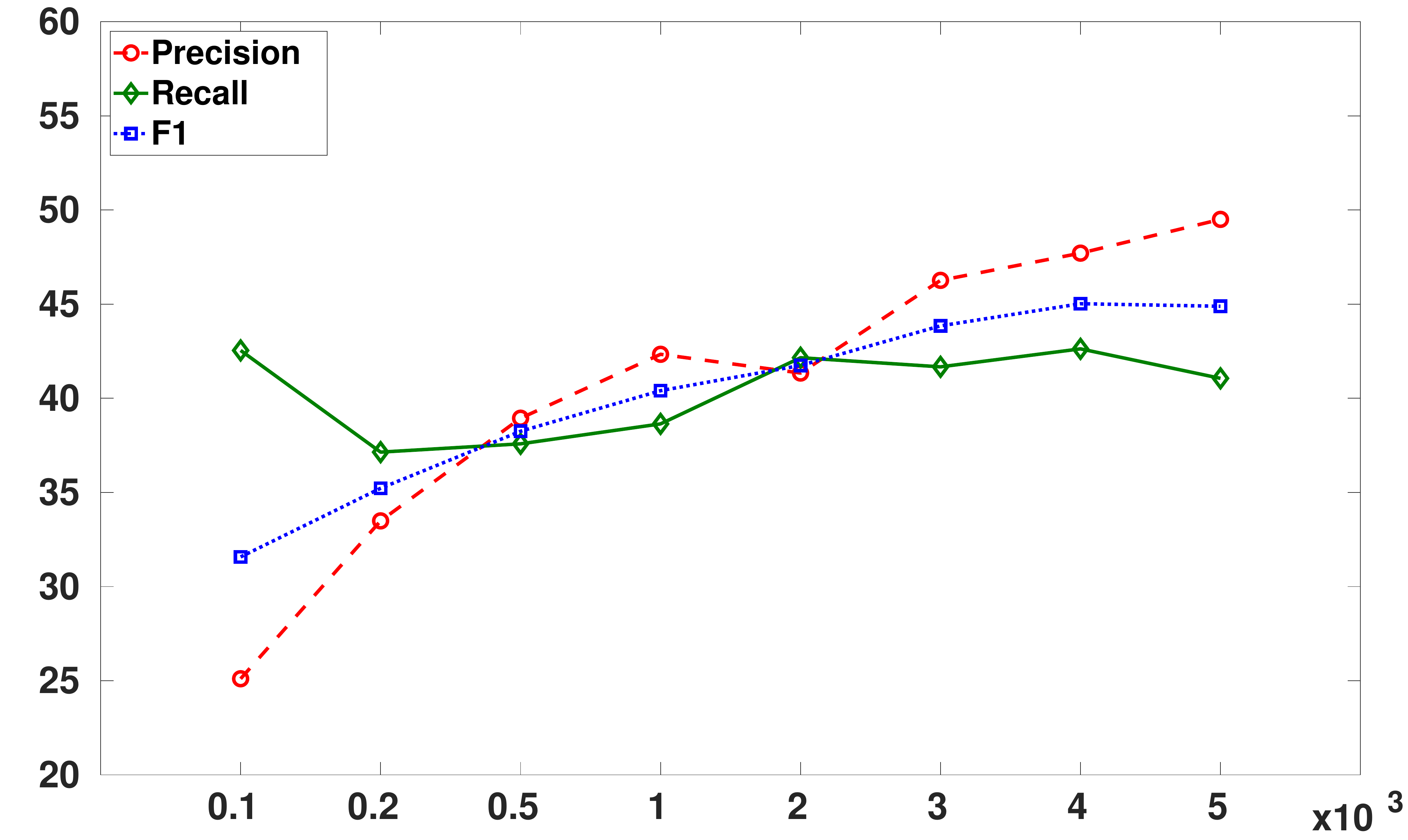}} &  \makecell{\includegraphics[width=.42\linewidth,height=0.30\linewidth]{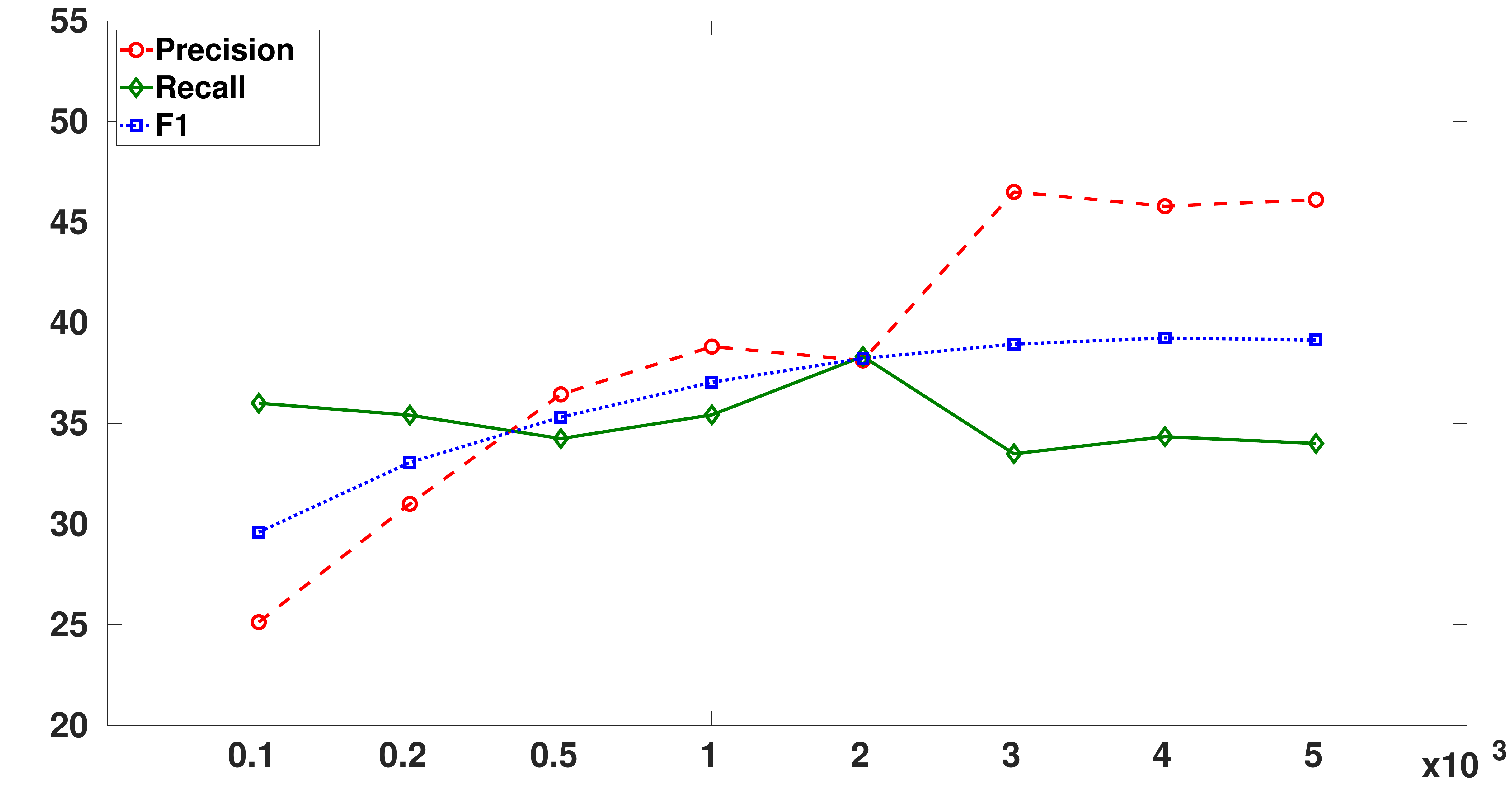}}  
\\
  \textbf{\rotatebox[origin=c]{90}{EfficientNet-B7}} &  \makecell{\includegraphics[width=.42\linewidth,height=0.30\linewidth]{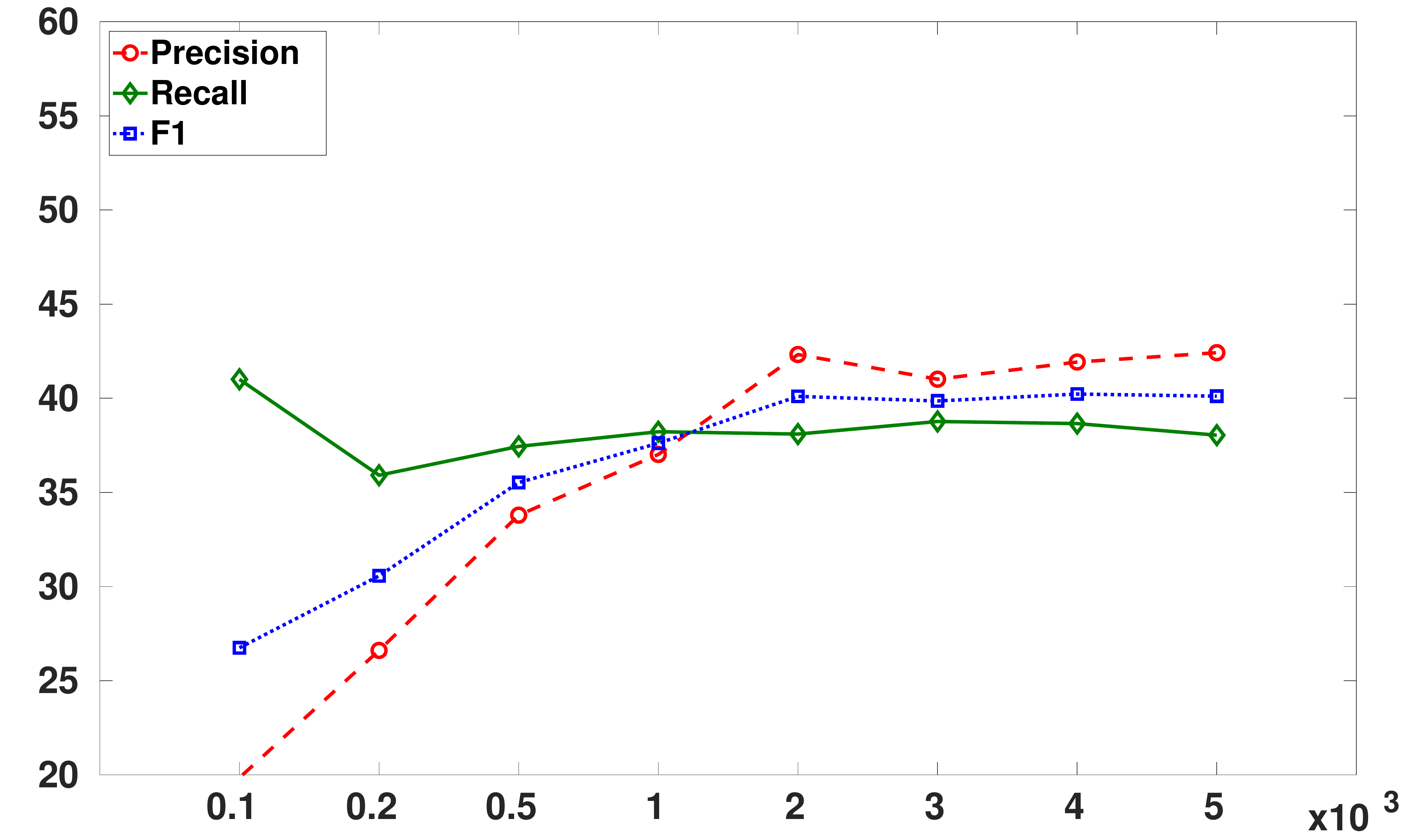}} &  \makecell{\includegraphics[width=.42\linewidth,height=0.30\linewidth]{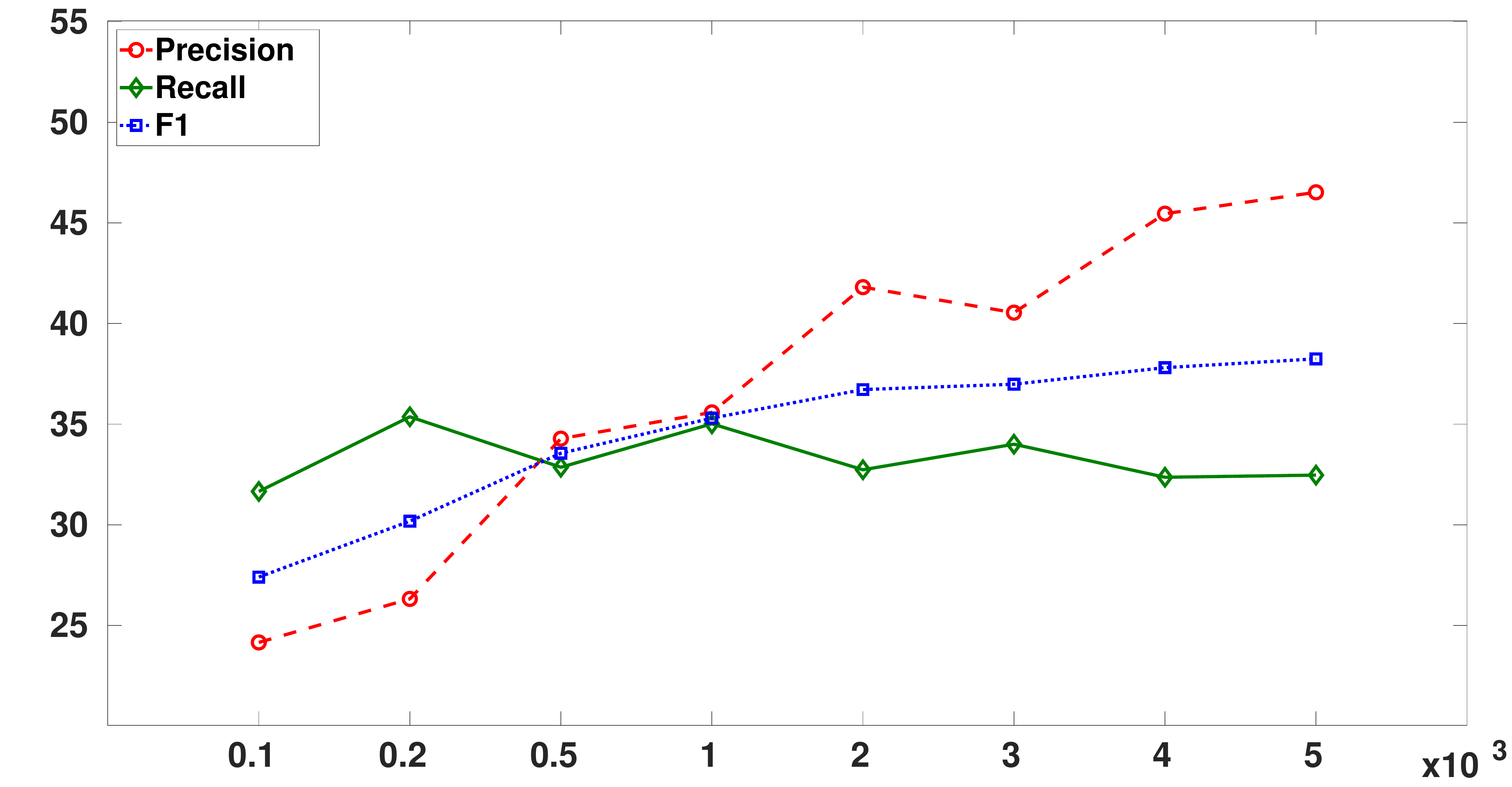}} 
 \\
\end{tabular}
\caption{Precision, Recall, and $F_1$ for IAPRTC-12 and ESP-GAME w.r.t the different number of prototypes. On each diagram, the x-axis shows the number of prototypes (in kilo) and the y-axis demonstrates the measures (in percent).}
\label{fig_dictsize1}
\end{figure}

\begin{figure}
\begin{tabular}{ccc}
  \textbf{} &  \textbf{FLICKR-60K} &  \textbf{FLICKR-125K}  \\
  \textbf{\rotatebox[origin=c]{90}{DenseNet-161}} & \makecell{\includegraphics[width=.42\linewidth,height=0.30\linewidth]{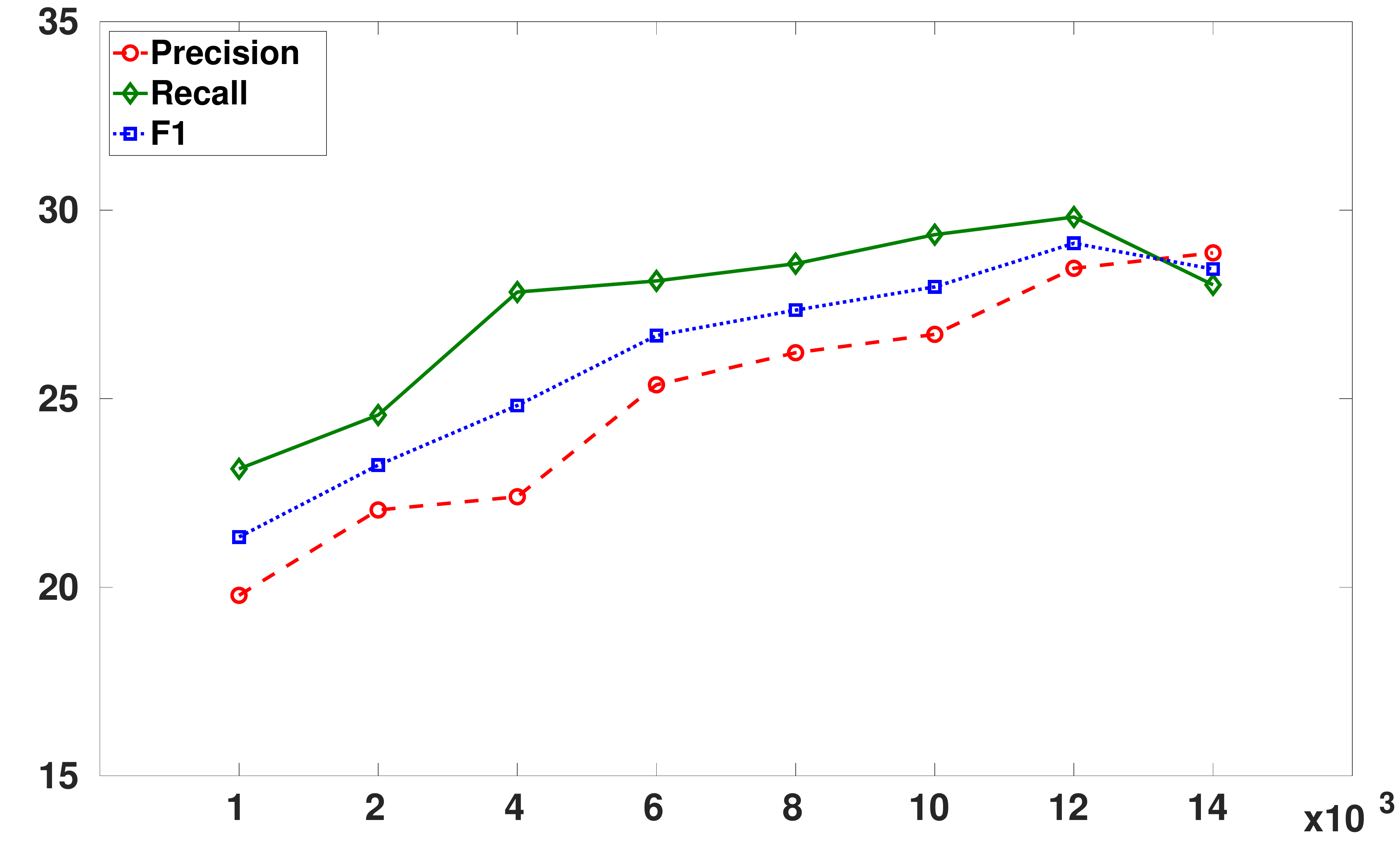}} &  \makecell{\includegraphics[width=.42\linewidth,height=0.30\linewidth]{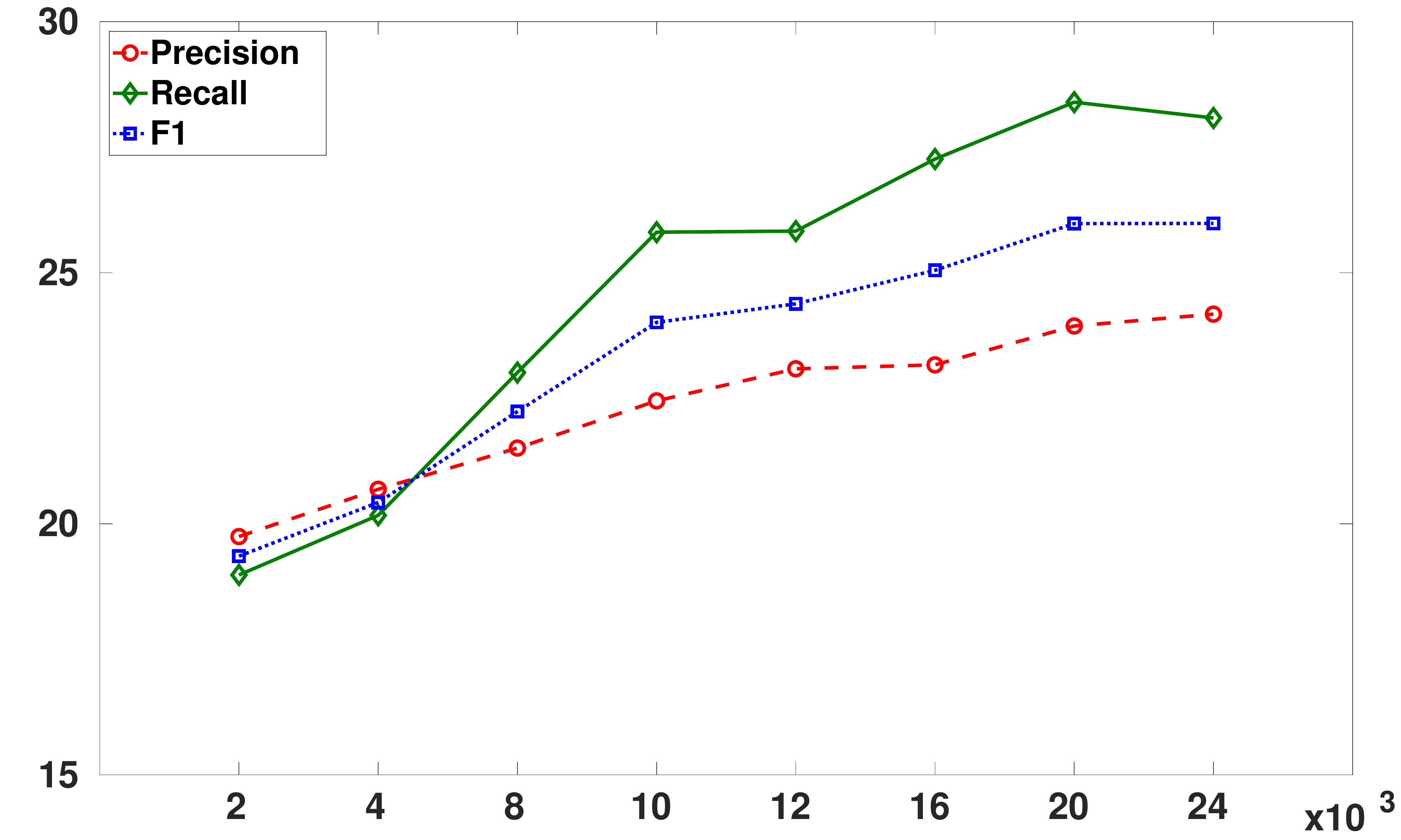}}  \\
  \textbf{\rotatebox[origin=c]{90}{ResNet-101}} &  \makecell{\includegraphics[width=.42\linewidth,height=0.30\linewidth]{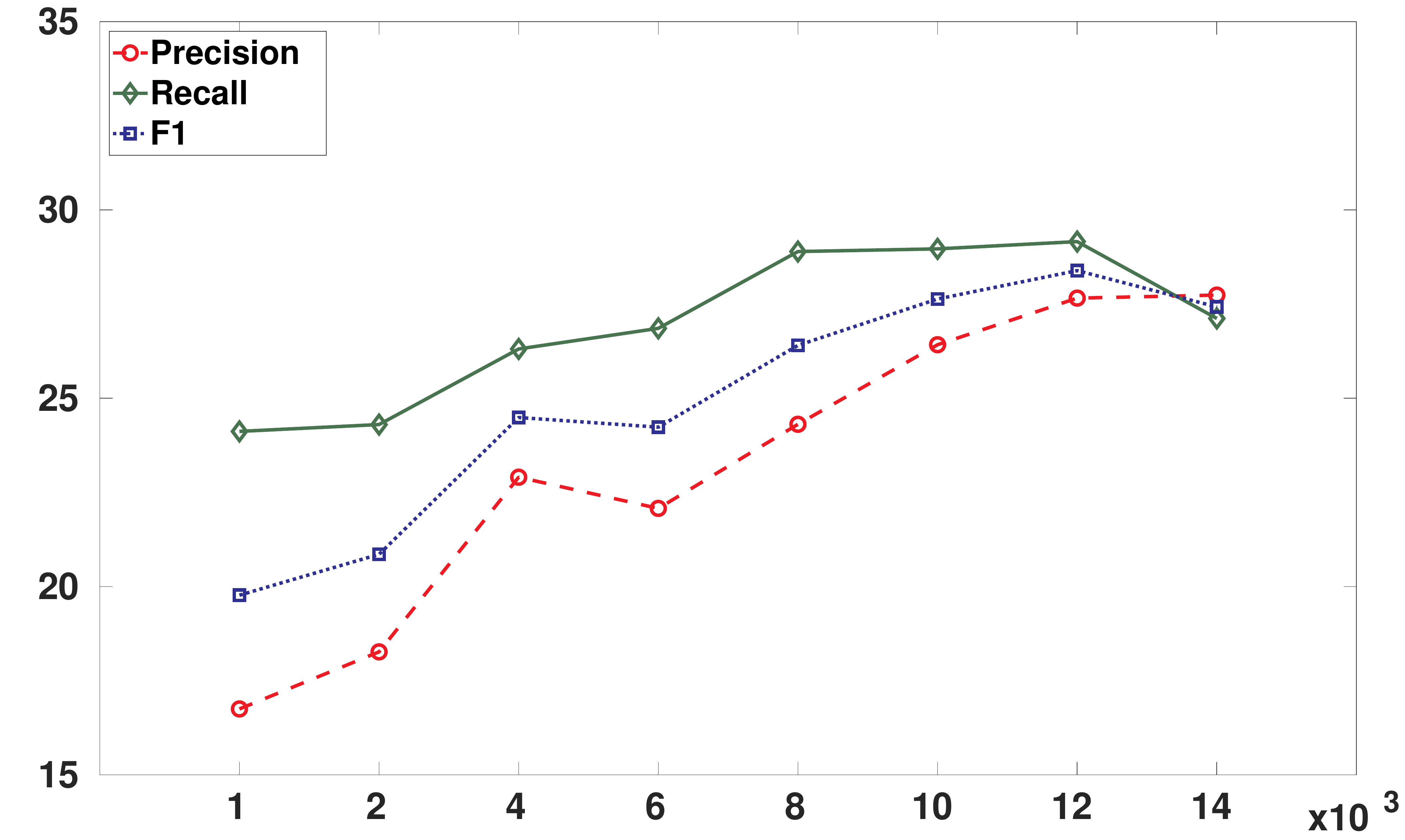}} &  \makecell{\includegraphics[width=.42\linewidth,height=0.30\linewidth]{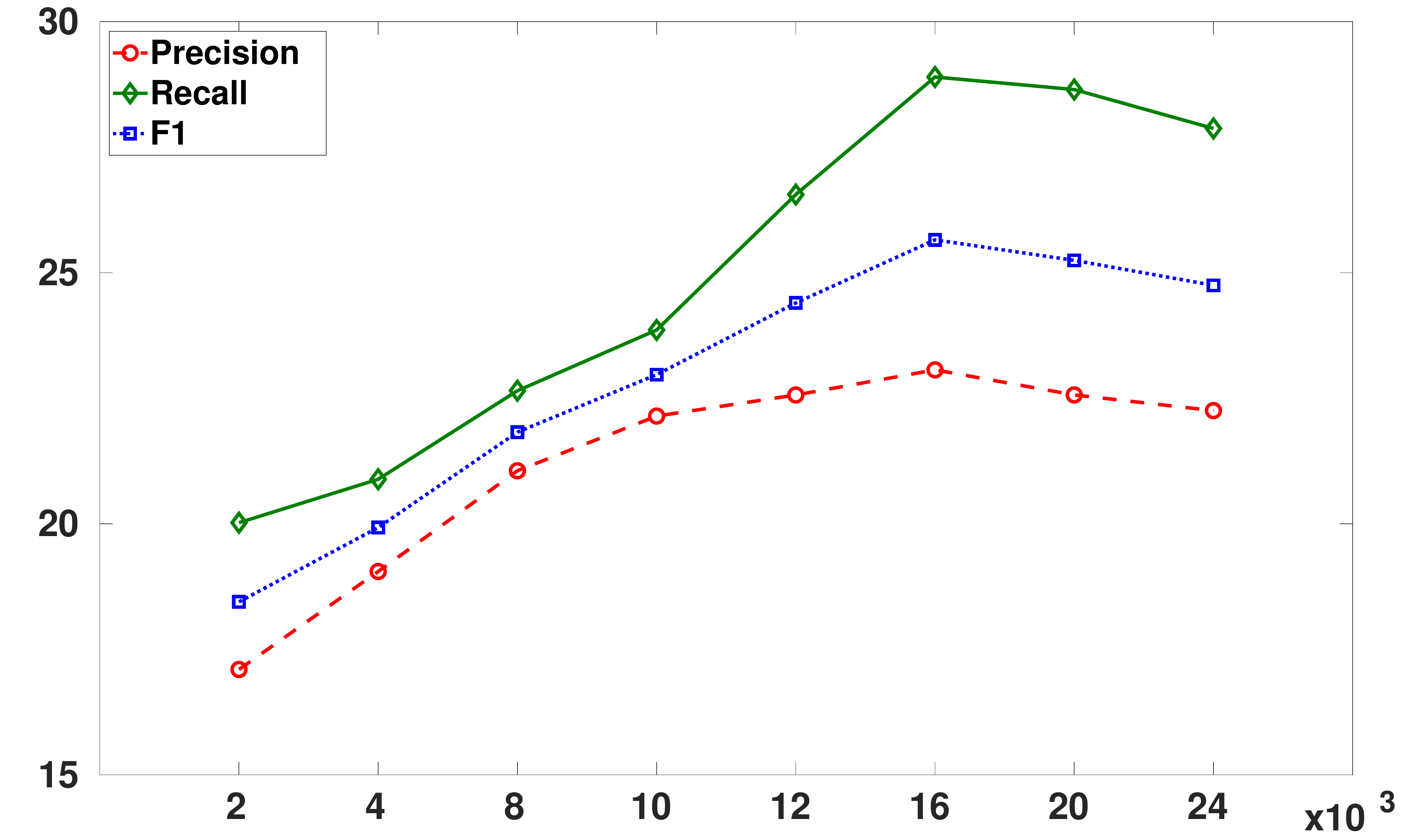}}  \\
%  \rotatebox{90}{\textbf{VggNet-16}} &  \makecell{\includegraphics[width=.42\linewidth]{FLICKR60K_Vgg16_Constant.pdf}} & 
\textbf{\rotatebox[origin=c]{90}{VggNet-16}} &  \makecell{\includegraphics[width=.42\linewidth,height=0.30\linewidth]{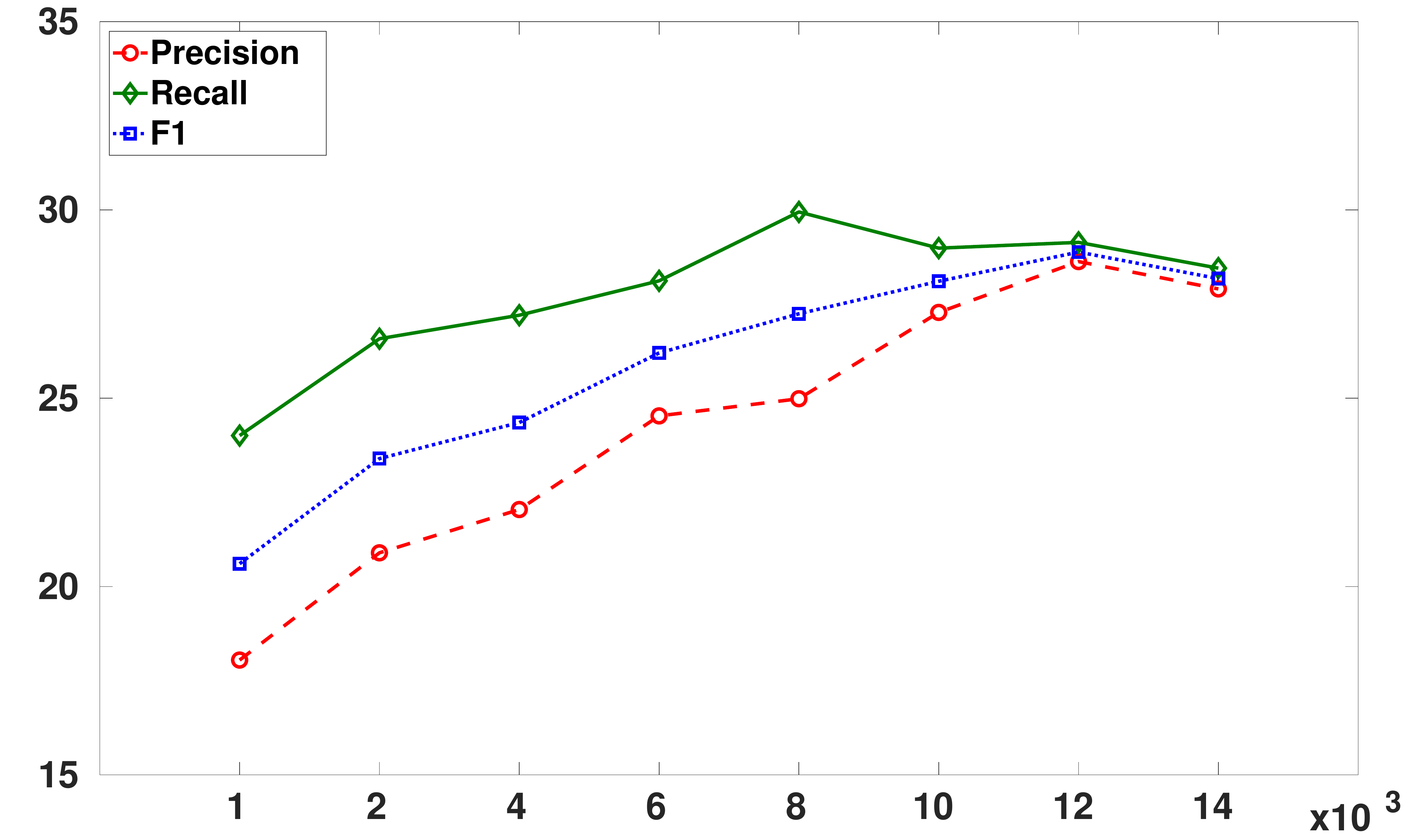}} &  \makecell{\includegraphics[width=.42\linewidth,height=0.30\linewidth]{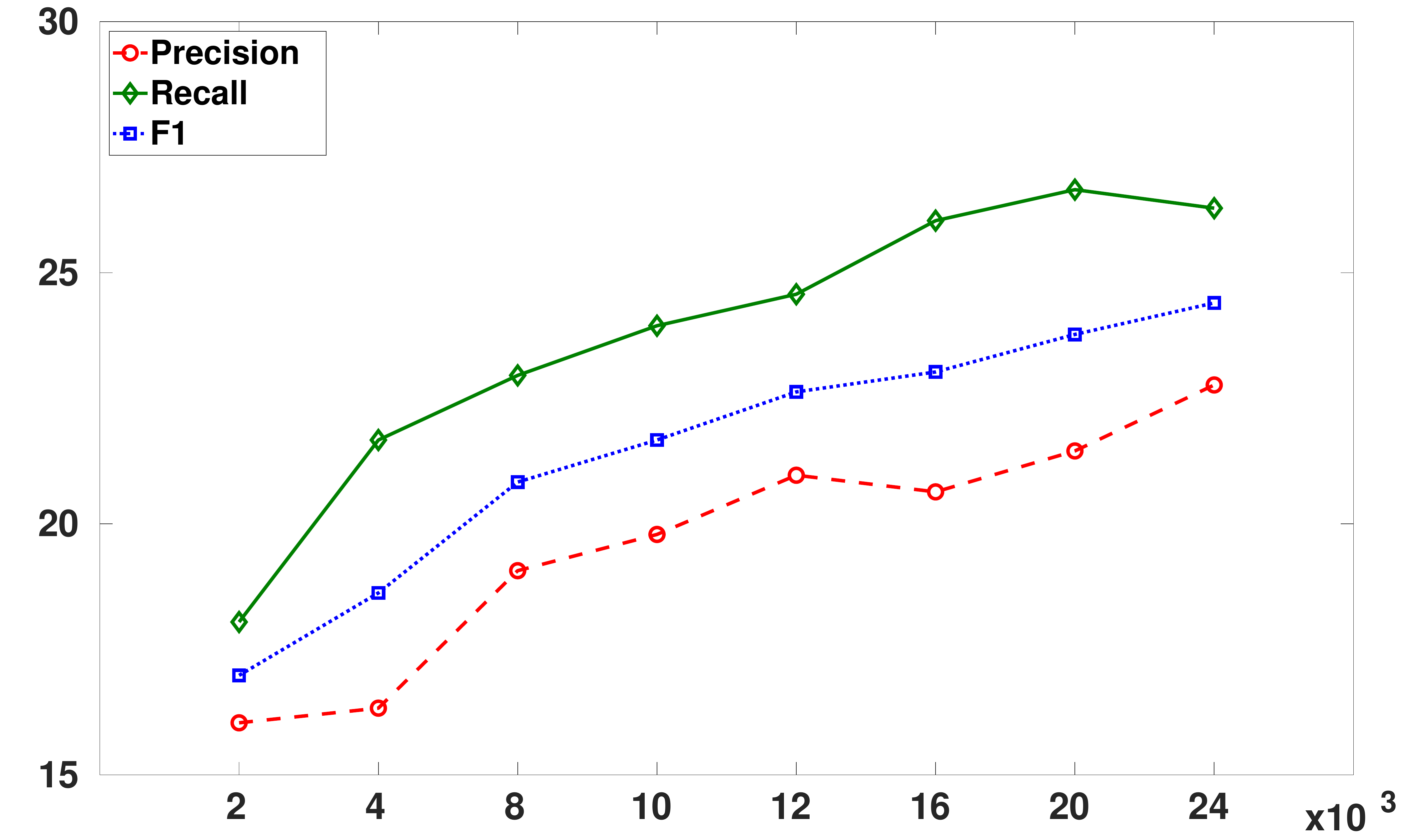}}  \\
\textbf{\rotatebox[origin=c]{90}{EfficientNet-B7}} &  \makecell{\includegraphics[width=.42\linewidth,height=0.30\linewidth]{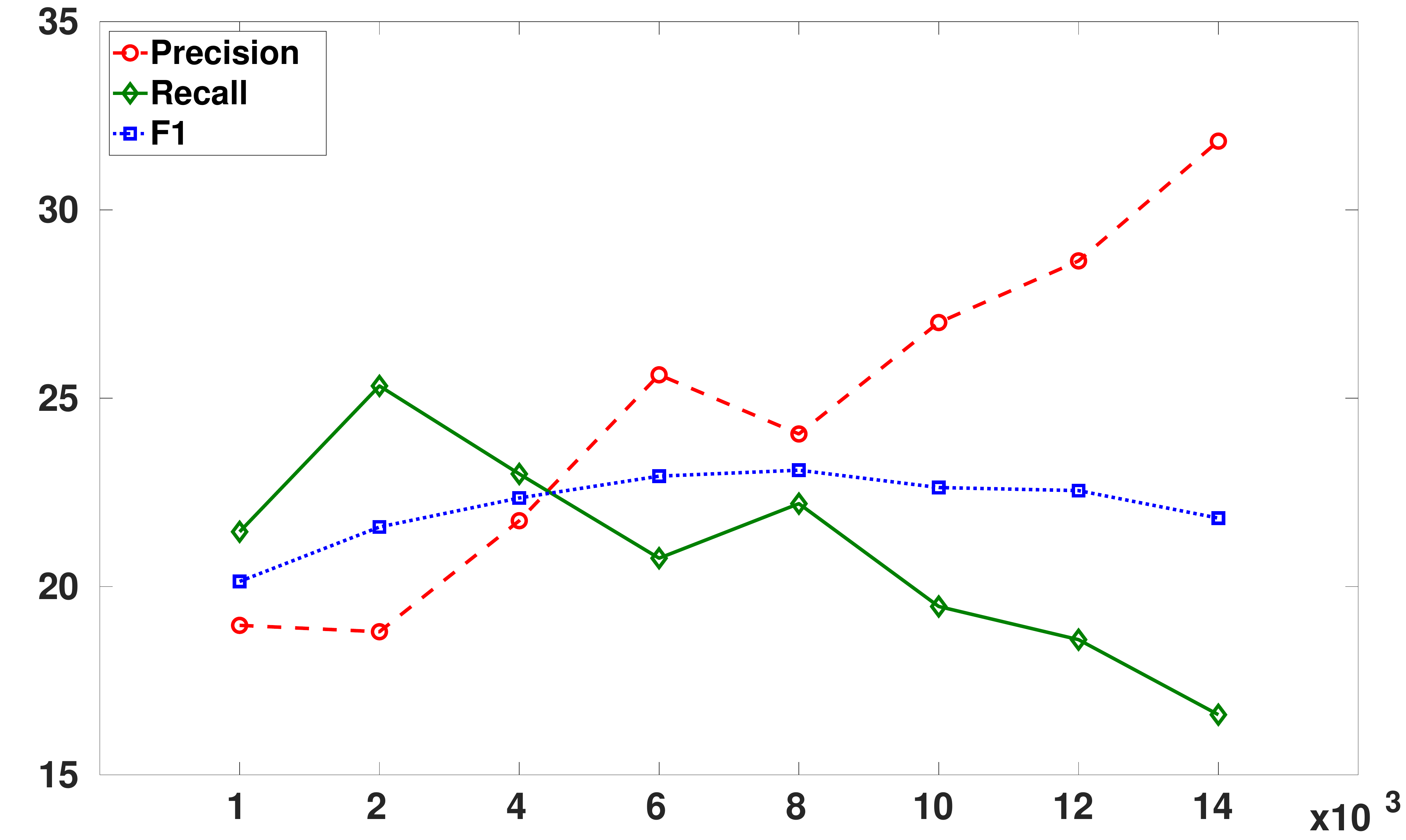}} &  \makecell{\includegraphics[width=.42\linewidth,height=0.30\linewidth]{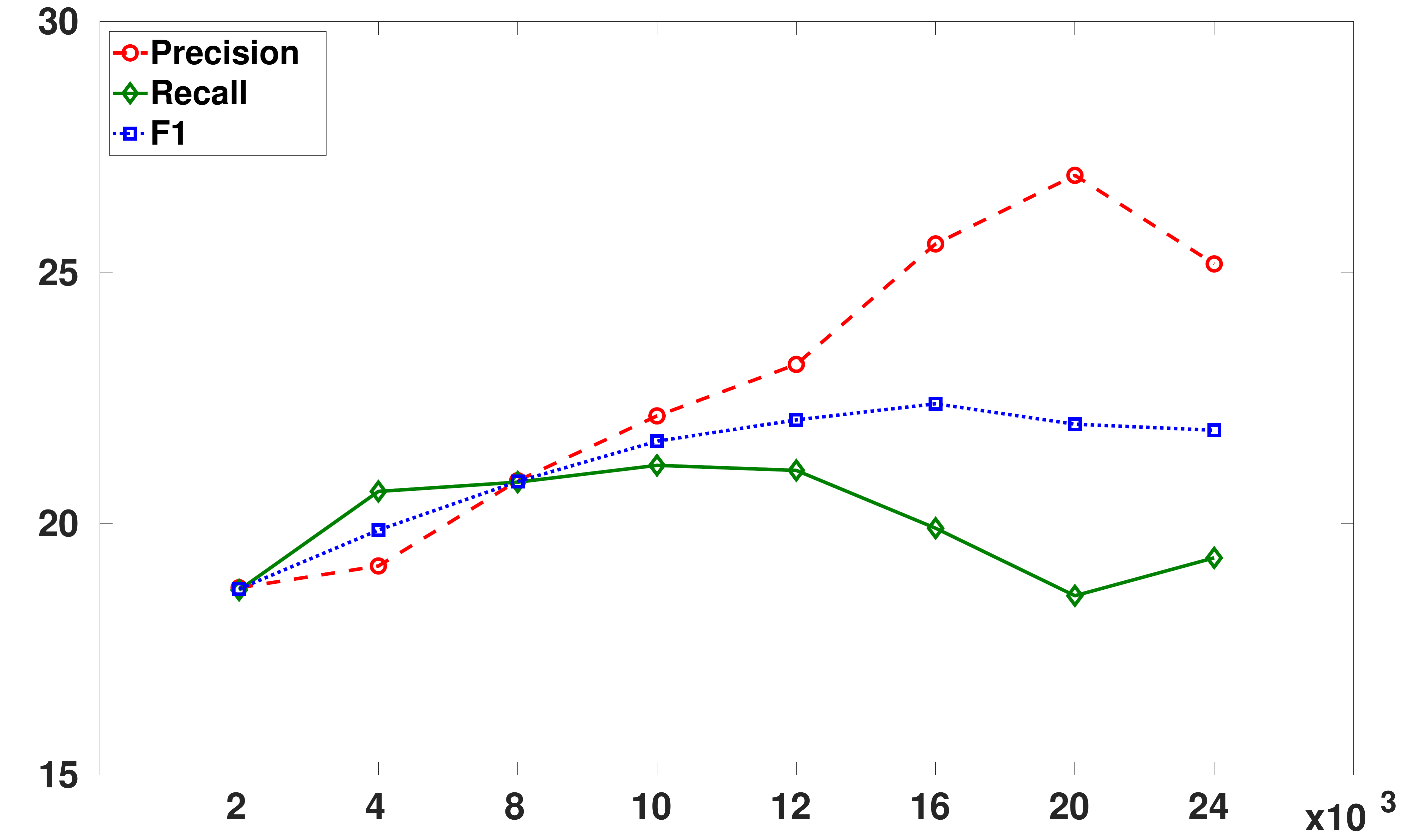}}  \\
\end{tabular}
\caption{Precision, Recall, and $F_1$ for FLICKR-60K and FLICKR-125K w.r.t the different number of prototypes. On each diagram, the x-axis shows the number of prototypes (in kilo) and the y-axis demonstrates the measures (in percent).}
\label{fig_dictsize2}
\end{figure}

Figures \mbox{\ref{fig_dictsize1}} and \mbox{\ref{fig_dictsize2}} illustrate the impact of dictionary size on precision, recall, and $F_1$ measures for the proposed MCDL algorithm applied to three different datasets. Starting from the lowest dictionary size for IAPRTC-12, which is a small value of 100, $F_1$ measure for all three features has an admissible value of over 30 percent. This indicates that the learned prototypes using MCDL are comprehensive candidates for training images. As we will discuss in the next section, this effectiveness originates from the marginalized loss function and ${\ell}_1-norm$. Although the increased dictionary size for IAPRTC-12 has improved the annotation measures, the increase rate is slower by exceeding the dictionary size of 3000. The best results are achieved through the dictionary size of 4000 in IAPRTC-12 dataset. For ESP-GAME also the trend is almost incremental. However, the best results are in the dictionary size of 4000. The overall initial $F_1$ for FLICKR-60K is around 20 percent and the best dictionary size for all feature types is 12000, except for EfficientNet-B7 which is a=8000. By increasing the dictionary size to above 12000 and 8000, the $F_1$ measure has decreased for this dataset. A plausible reason for such decreases in $F_1$ measure is the overfitting phenomenon for semantic dictionary, which is common when the number of parameters is increased, and in our case, the can be biased to For FLICKR-125K, $F_1$ has gradually increased before reaching the best value of around 16000 and 20000 prototypes.  As it can be seen, EfficientNet-B7 has poor results versus other networks. Experimentally, if a network provides features which is more linear separable, MCDL can perform better. \\

%\\
Figures \ref{fig_dictsize1} and \ref{fig_dictsize2} show that MCDL can achieve admissible results with a small number of prototypes. The best dictionary size is about 4000 for IAPRTC-12 and ESP-GAME, and 12000, and 20000 for two FLICKR-60K and 125K. These dictionary sizes are around 20 percent of the training data size, indicating the efficiency of our approach for summarizing large datasets to a limited number of prototypes.
%%%%%%%%%% ANALYSIS OF THE OBJECTIVE FUNCTION %%%%%%%%%%%
\subsection{Analysis of The Objective Function}
\label{sec_analysis_obj}
In this section, the effectiveness of different stages of the proposed method is assessed. As the first baseline, we have eliminated coupled learning in MCDL by examining unsupervised dictionary learning called UDL. In this baseline, visual dictionary is first learned the same as Step \hyperref[itm:step21]{2.1} of Algorithm \ref{algo_learning}. Then, semantic labels of learned prototypes are obtained using Step \hyperref[itm:step22]{2.2} of Algorithm \ref{algo_learning}. Furthermore, to study the importance of marginalized loss function and ${\ell}_1$ regularization in MCDL, we have examined another baseline, named Coupled Dictionary Learning (CDL), where hinge loss has been replaced with squared loss function and ${\ell}_1$ regularization is omitted.
\begin{figure}
\begin{tabular}{cc}
\makecell{\textbf{IAPRTC-12}} &  {\textbf{ESP-GAME}}  \\
\makecell{\includegraphics[width=.46\linewidth]{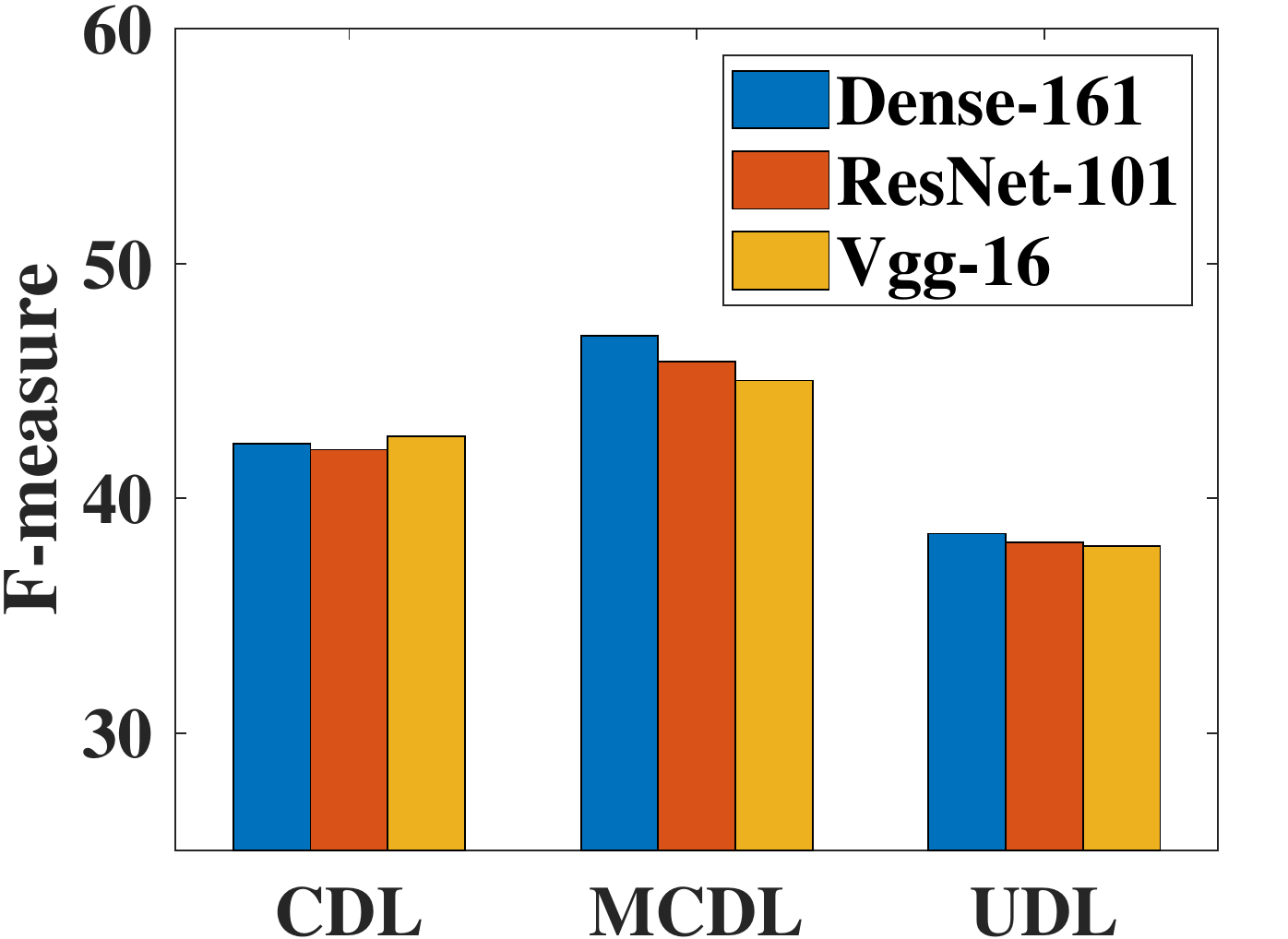}} &  \makecell{\includegraphics[width=.46\linewidth]{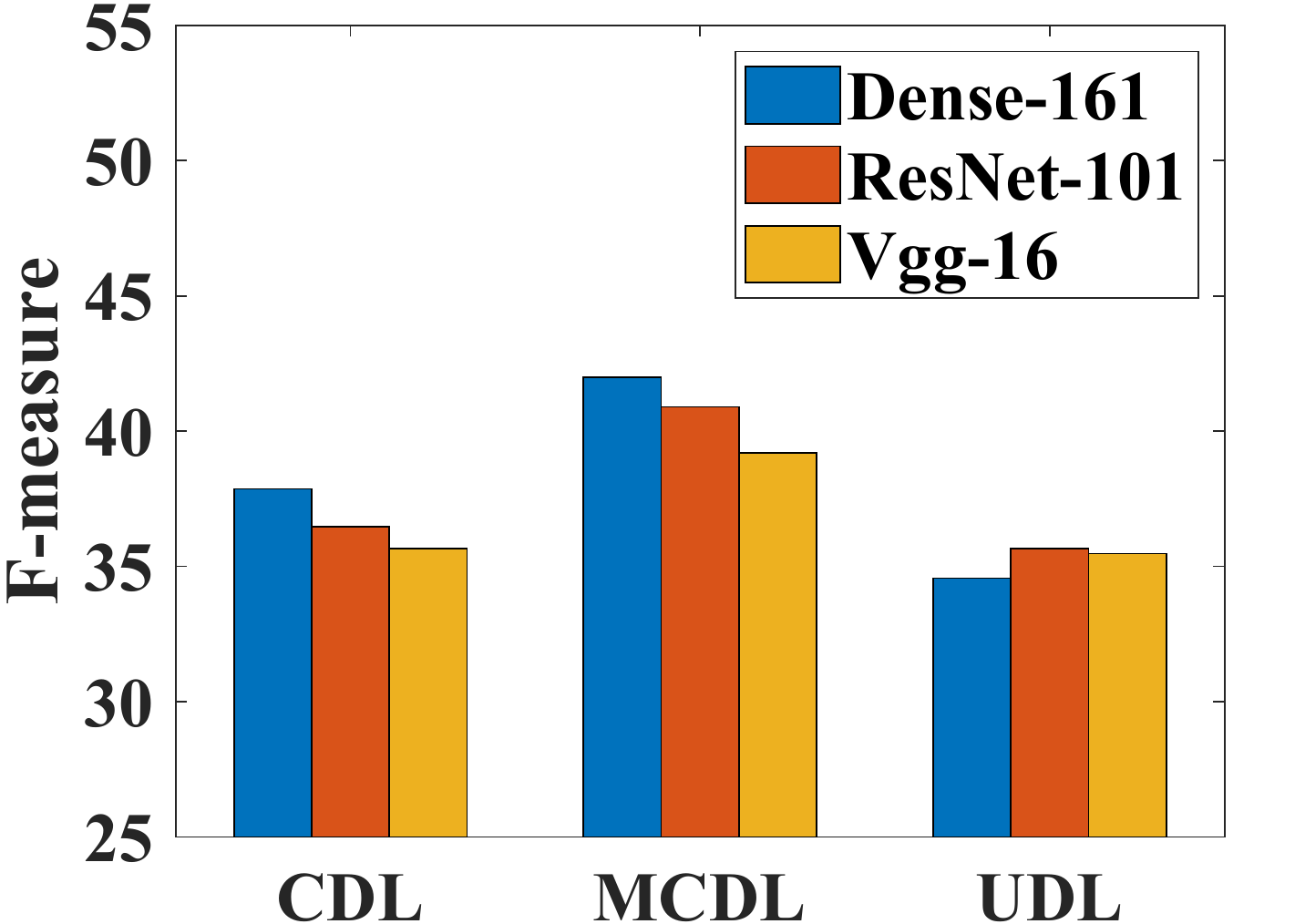}}  \\ 
\makecell{\textbf{FLICKR-60K}} &  {\textbf{FLICKR-125K}} \\
\makecell{\includegraphics[width=.46\linewidth]{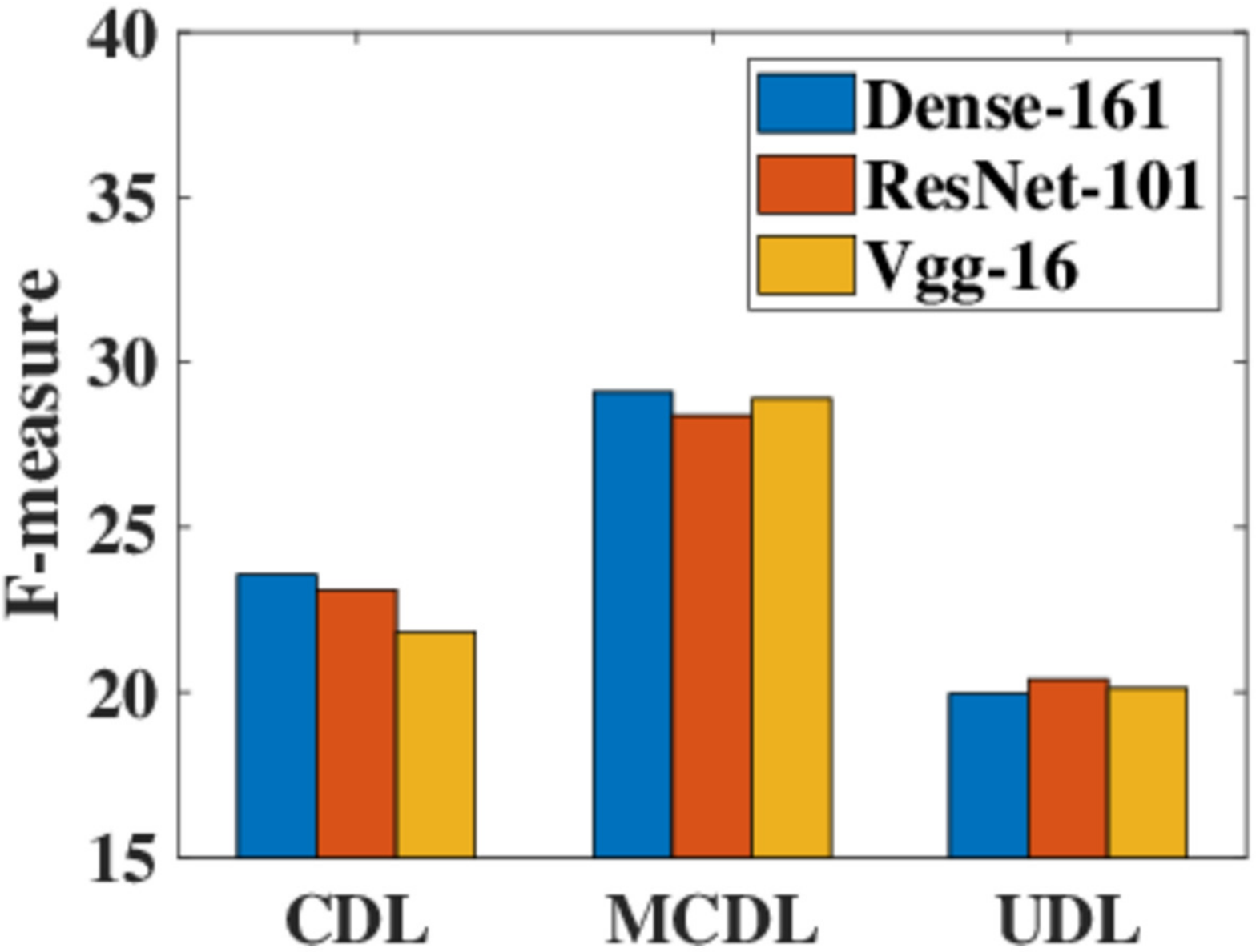}} &  \makecell{\includegraphics[width=.46\linewidth]{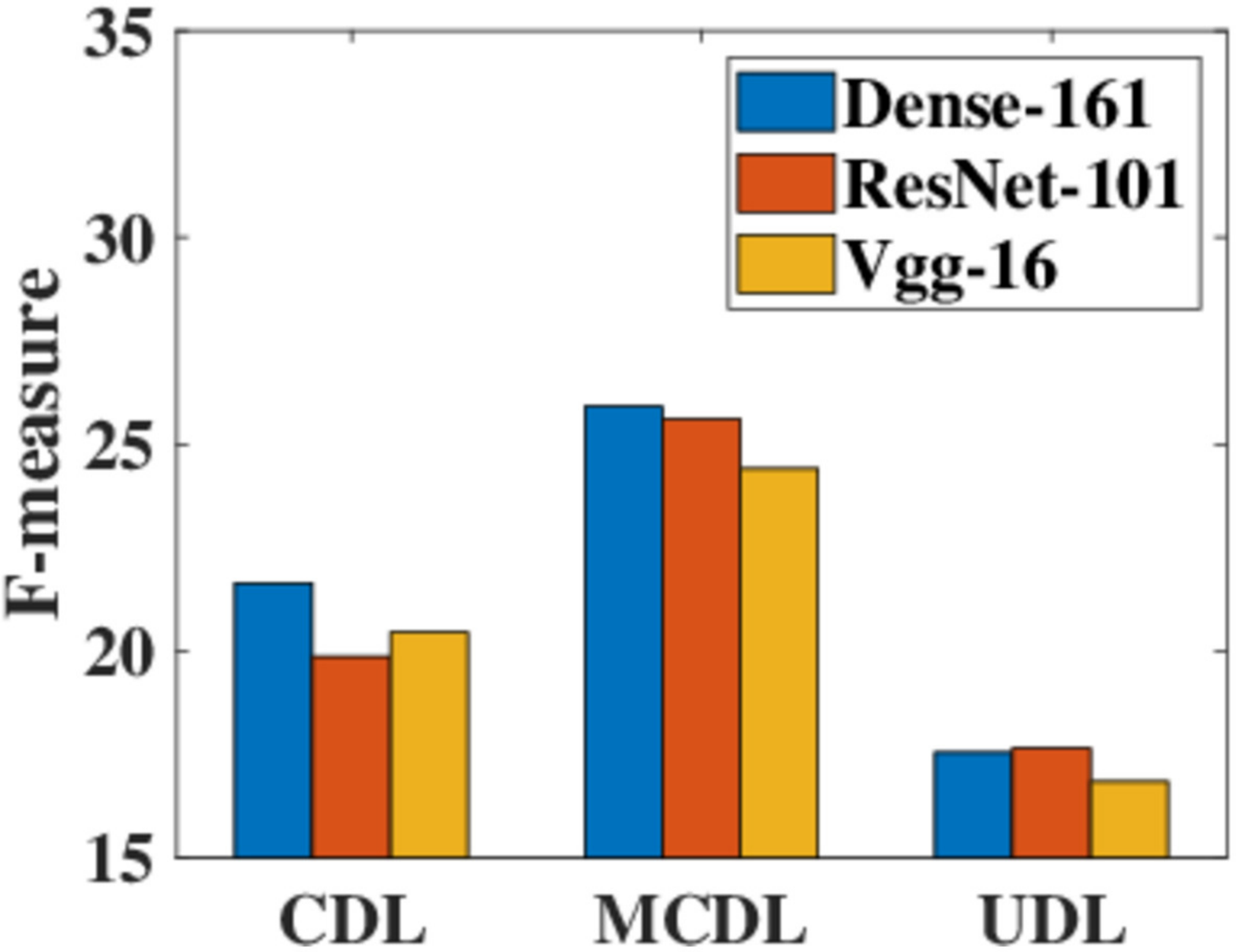}}  \\
\end{tabular}
\caption{$F_1$ measure for CDL and UDL versus the proposed MCDL.}
\label{fig_coupledimpact}
\end{figure}
\\
Results in Figure \ref{fig_coupledimpact} are provided based on the best dictionary size mentioned in the previous section. It illustrates that CDL method (supervised version of UDL) has better performance with significant differences in large-scale datasets. It is noticeable that the marginalized hinge loss used in MCDL method has raised the generalization of the prototypes significantly. Moreover, CDL method uses squared loss function instead of the marginalized loss function suggested in the MCDL. Although least square loss function is appropriate for visual modality, it yields to biased prototypes for imbalanced zero-one labels, as it can be investigated from Figure \ref{fig_margin}. As figure shows, most of the scores are concentrated around zero (labels are considered 0 and 5).
%%%%%%%%%% SACLABILITY ANALYSIS AND COMPARISON %%%%%%%%%%%

\subsection{Scalability Analysis and Comparison}

This section focuses on scalability analysis based on two important criteria, including performance and run-time. We have considered one of the most popular and state-of-the-art similarity-based approaches, 2PKNN, as the main baseline method. To reach a fair comparison, we have used the same feature vectors mentioned in the previous sections for both 2PKNN and our approach.\\
\textbf{Feature Analysis}\\
Table \ref{tab_compare} reveals that in IAPRTC-12 dataset, MCDL considerably outperforms the baseline method, 2PKNN, in all three feature types. The improvement is about 9.8 percent for feature type of DenseNet-161. For the second dataset, ESP-GAME, the results have also been improved through MCDL and the most considerable improvement is 4.1 percent. On the other hand, in FLICKR-60K, the most egregious progress is related to VggNet-16, where it reaches to 29 percent in MCDL from 26 percent in 2PKNN. Finally, the results for 125K version of FLICKR are improved in ResNet-101 and VggNet-16. The reason for the better performance of MCDL on IAPRTC-12 and ESP-GAME rather than two FLICKR datasets is that there is considerable redundancy in the former datasets, as can be investigated from Figure \ref{fig_dictsize1} and \ref{fig_dictsize2}. Indeed, when we retrieve a fixed number of similar images to annotate a query image, the retrieved images may be highly correlated in their visual contents and semantic labels. However, MCDL tries to reconstruct query images based on various prototypes and thus the redundancy will be reduced.\\
%\begin{table}
%\caption{$F_1$ comparison between MCDL and 2PKNN.}
%\label{tab_compare}       % Give a unique label
%\begin{tabular}{ccccc}
%\hline\noalign{\smallskip}
% & & \textbf{DenseNet-161} & \textbf{ResNet-101} & \textbf{VggNet-16} \\
%\multirow{2}{*}{\textbf{IAPRTC-12}} & \textit{2PKNN} & 37.1 & 37.3 & %37.6 \\
% & \textit{MCDL} & 46.9 & 45.8 & 45.0 \\
% \noalign{\smallskip}\hline\noalign{\smallskip} 
%\multirow{2}{*}{\textbf{ESP-GAME}} & \textit{2PKNN} & 37.9 & 37.0 & %36.4 \\
% & \textit{MCDL} & 42.0 & 40.9 & 39.2 \\
% \noalign{\smallskip}\hline\noalign{\smallskip} 
%\multirow{2}{*}{\textbf{FLICKR-60K}} & \textit{2PKNN} & 28.9 & 28.5 & %26.4 \\
% & \textit{MCDL} & 29.1 & 28.4 & 28.9 \\
% \noalign{\smallskip}\hline\noalign{\smallskip} 
%\multirow{2}{*}{\textbf{FLICKR-125K}} & \textit{2PKNN} & 27.1 & 25.5 & %23.2 \\
% & \textit{MCDL} & 25.9 & 25.6 & 24.4 \\
%\end{tabular}
%\end{table}
\begin{table}
\caption{$F_1$ comparison between MCDL and 2PKNN (in percent).}
\label{tab_compare}       % Give a unique label
\begin{tabular}{cccc}
\hline\noalign{\smallskip}
\textit{DATASET} & \textit{FEATURE} & \textit{2PKNN} & \textit{MCDL}\\
\hline\noalign{\smallskip}
\multirow{3}{*}{\textbf{IAPRTC-12}} & \textbf{DenseNet-161} & 37.1 & 46.9\\
 & \textbf{ResNet-101} & 37.3 & 45.8\\
 & \textbf{VggNet-16} & 37.6 & 45.0\\
 & \textbf{EfficientNet-B7} & 38.0 & 40.2 \\ %0.4733    0.2361    0.3150  210.0000
 \noalign{\smallskip}\hline\noalign{\smallskip} 
\multirow{3}{*}{\textbf{ESP-GAME}} & \textbf{DenseNet-161} & 37.9 & 42.0\\
 & \textbf{ResNet-101} & 37.0 & 40.9\\
 & \textbf{VggNet-16} & 36.4 & 39.2\\
 & \textbf{EfficientNet-B7} & 36.4 & 38.2\\ %0.4733    0.2361    0.3150  210.0000
 \noalign{\smallskip}\hline\noalign{\smallskip} 
\multirow{3}{*}{\textbf{FLICKR-60K}} & \textbf{DenseNet-161} & 28.9 & 29.1\\
 & \textbf{ResNet-101} & 28.5 & 28.4\\
 & \textbf{VggNet-16} & 26.4 & 28.9\\
 & \textbf{EfficientNet-B7} & 20.5 & 23.0\\ %0.4733    0.2361    0.3150  210.0000
 \noalign{\smallskip}\hline\noalign{\smallskip} 
\multirow{3}{*}{\textbf{FLICKR-125K}} & \textbf{DenseNet-161} & 27.1 & 25.9\\
 & \textbf{ResNet-101} & 25.5 & 25.6\\
 & \textbf{VggNet-16} & 23.2 & 24.4\\
 & \textbf{EfficientNet-B7} & 18.5 & 22.4\\ %0.4733    0.2361    0.3150  210.0000
\noalign{\smallskip}\hline
\end{tabular}
\end{table}
\textbf{Annotation Time}\\
Table \ref{tab_time} compares the annotation time of MCDL against the baseline. The experiments are conducted on a PC with an Intel (R) Core (TM) i7-6700 HQ 3.1 GHz CPU, and 16G RAM, in MATLAB environment. Furthermore, the annotation time is averaged over all the test images. While 2PKNN needs a tremendous number of peer to peer comparisons for finding the most similar images per each label, MCDL needs a tiny proportion of 2PKNN time to annotate an input image. For IAPRTC-12 and ESP-GAME with roughly 20000 images, labeling a new image takes over 25 milliseconds using the 2PKNN method. This measure is sharply declined by MCDL to under 1.5 milliseconds. To sum up, the information presented in Tables \ref{tab_compare} and \ref{tab_time} implies that not only the scalability is acquired, but also the performance of MCDL is improved.\\
\begin{table}
\caption{The average annotation time (in millisecond) for input images in MCDL compared to 2PKNN, using DenseNet-161 feature vector. The third column shows the percentage of the reduction in annotation time using MCDL method.}
\label{tab_time}
\begin{tabular}{cccc}
\hline\noalign{\smallskip}
 & \textbf{2PKNN} & \textbf{MCDL}  & \textbf{Reduction}\\
\noalign{\smallskip}\hline\noalign{\smallskip}
\textbf{IAPRTC-12} & 27.5 & 1.5 & 94.5\%\\
\textbf{ESP-GAME} & 25.4 & 1.2 & 95.2\%\\
\textbf{FLICKR-60K} & 57 & 1.8 & 96.8\%\\
\textbf{FLICKR-125K} & 390 & 10 & 97.4\%\\
\noalign{\smallskip}\hline
\end{tabular}
\end{table}
\textbf{Performance Analysis}\\
To compare the performance of our approach against the other methods, we provide Table \mbox{\ref{tab_compareall}}.  It is necessary to mention that there are three reports for 2PKNN in this table. The first, 2PKNN (SD), is its performance on traditional standard features. The second utilizes the same features with a metric learning algorithm. In 2PKNN (CNN), we fed it our CNN-based features to make a fair comparison by our approach. The initial impression of this table is that CNN-based features could provide better performances in comparison with the standard features (i.e., color, histogram, shape, sift, etc.). A meticulous glance at the precision and recall values reveals that there is a considerable variance between them in almost all the methods. Fortunately, this is not true in our method. The reason is that, against the others, MCDL assigns different number of labels to any input sample based on its scores. The other methods take a fixed number of labels for annotation, which is mainly less than the required. Therefore while the precision improves, the recall does not. This fact originates from the increase of the number of false-negatives. The reason for a good trade-off between precision and recall in MCDL is that our technique utilizes a marginalized approach on scores. Table \mbox{\ref{tab_compareall}} demonstrates that in both datasets, IAPRTC-12 and ESP-GAME, the proposed MCDL method could achieve the highest $F_1$ scores, 47 percent for IAPRTC-12 and 42 percent for ESP-GAME. Looking at the precision values of Table \mbox{\ref{tab_compareall}} depicts that among the standard-feature-based methods MLDL and ML-based 2PKNN can provide better results than our method. This is also true for 2PKNN with CNN features. On the other hand, the recall value for MCDL is significantly higher than all the methods, and this is the reason that our method could pick the best $F_1$ score. \\
One more point, in Table \ref{tab_2pknn} we compare our approach and baseline method on FLICKR-60K and FLICKR-125K datasets where it is obvious that those $F_1$ scores are significantly close to each other. The importance of this subject is clarified when we note that our approach reaches this $F_1$ score by replacing all training images with a few prototypes (20000 prototypes instead of 124840 images of FLICKR-125K). This property leads to a considerable reduction in the annotation time, as presented in Table \ref{tab_time}.\\

\begin{table}
\caption{Precision, Recall, and $F_1$ comparison of different methods on IAPRTC-12 and ESP-GAME datasets (in percent) using Dense-161 feature .}
\label{tab_compareall}
\begin{center}
\begin{tabular}{ccccc|ccc}
\hline\noalign{\smallskip}
 \multicolumn{2}{c}{\multirow{2}{*}{\textbf{Method}}} & \multicolumn{3}{c}{\textbf{IAPRTC-12}} & \multicolumn{3}{c}{\textbf{ESP-GAME}} \\
\cline{3-5}\cline{6-8}
& & \textit{P} & \textit{R} & \textit{$F_1$} & \textit{P} & \textit{R} & \textit{$F_1$} \\
 \noalign{\smallskip}\hline\noalign{\smallskip}
\multirow{8}{*}{\rotatebox{90}{Standard Feature}}&\textit{ML \cite{guillaumin2009tagprop}} & 48 & 25 & 33 & 49 & 20 & 28 \\
&\textit{$\sigma$ML \cite{guillaumin2009tagprop}} & 46 & 35 & 40 & 39 & 27 & 32 \\
&\textit{Fast Tag \cite{03-07-chen13}} & 47 & 26 & 34 & 46 & 22 & 30 \\
&\textit{KSVM-VT \cite{B14-verma13}} & 47 & 29 & 36 & 33 & 32 & 33 \\
&\textit{MLDL \cite{000030-jing16}} & 56 & 40 & 47  & 56 & 31 & 40 \\
&\textit{2PKNN (SD) \cite{verma2017image}} & 49 & 32 & 39 & 51 & 23 & 32 \\
&\textit{2PKNN (ML) \cite{verma2017image}} & 54 & 37 & 44 & 53 & 27 & 36 \\
&\textit{Mvg-NMF \cite{rad2018multi}}  & 47 & 40 & 43 & 41 & 33 & 37 \\
\hline
\multirow{8}{*}{\rotatebox{90}{CNN Feature}}&\textit{MVSAE \cite{03-09-yang15}} & 43 & 38 & 40 & 47 & 28 & 34 \\
&\textit{CCA-KNN \cite{03-06-murthy15}} & 45 & 38 & 41 & 46 & 36 & 41 \\
&\textit{RPLRF \cite{li2017ranking}} & 48 & 29 & 36 & 43 & 27 & 34 \\
&\textit{AHL \cite{tang2019adaptive}} & 47 & 35 & 40 & 46 & 23 & 31 \\
&\textit{SEM \cite{ma2019cnn}} & 41 & 39 & 40 & 38 & 42 & 40 \\
&\textit{VLAD \cite{chen2020image}} & 46 & 33 & 38 & 44 & 33 & 38 \\
&\textit{2PKNN (CNN)} & 51 & 29 & 37 & 50 & 31 & 38 \\
&\textit{\textbf{MCDL}} & \textbf{49} & \textbf{45} & \textbf{47} & \textbf{46} & \textbf{39} & \textbf{42} \\
\noalign{\smallskip}\hline
\end{tabular}
\end{center}
\end{table}

\begin{table}
\caption{Precision, Recall and $F_1$ comparison of different methods on FLICKR-60K and FLICKR-125K datasets (in percent) using Dense-161 feature.}
\label{tab_2pknn}
\begin{center}
\begin{tabular}{ccccccc}
\hline\noalign{\smallskip}
\textbf{Method} & \qquad & \textbf{Dataset} &  \qquad & \textbf{Pre} & \textbf{Rec} & \textbf{$F_1$}\\
\noalign{\smallskip}\hline
\multirow{2}{*}{2PKNN} &  &  FLICKR-60K & & 34 & 25 & 29\\
 & & FLICKR-125K & & 32 & 24 & 27\\
 \noalign{\smallskip}\hline
\multirow{2}{*}{MCDL} & & FLICKR-60K & & 28 & 30 & 29\\
 & & FLICKR-125K & & 24 & 28 & 26\\
\noalign{\smallskip}\hline
\end{tabular}
\end{center}
\end{table}
\textbf{Computational Complexity}\\

In the matter of computational complexity, our method outperforms the baseline, 2PKNN. To begin with, we first apply a dimensionality reduction on each input visual vector to convert it to a low-dimension vector of size $M$. 2PKNN includes the distance computation of the input image with all training samples which have the time complexity of $O(N \times M)$, followed by finding the $K_1$ most similar training images \mbox{\cite{verma2019diverse}} for each label with time complexity of $O (N \times N \times T)$. Note that if linear algorithms are used to find first $K_1$  nearest neighbors instead of calling sort in the original algorithm of 2PKNN, it will be of $O (N  \times K_1 \times T)$).  Therefore, the total time complexity of 2PKNN is \mbox{$O (N \times M $} \mbox{$+  N \times N \times T)$} . \\
On the other hand, our method needs to solve a sparse coding using
Lasso \mbox{\cite{efron2004least}} in the annotation stage. First, we need to compute the gram matrix ${{\mathbf{D}}^I}^{\top} {{\mathbf{D}}^I}$ over the leaned dictionary, which can be pre-computed \mbox{\cite{mairal2010online}} beforehand and so does not impact the time complexity of MCDL. Then in the annotation stage,  ${{\mathbf{D}}^I}^{\top}x$ needs to be computed for a given input image with the time complexity of $O(K \times M)$. Then,  a  Cholesky-based algorithm with $O(K \times M^2)$ \mbox{\cite{efron2004least}} should be done to find the coefficient over ${{\mathbf{D}}^I}$. Thus, the overall complexity of the MCDL approach in the annotate step is  \mbox{$O(K \times M$} \mbox{$ + K \times M^2)$}. Since the number of prototypes is much less than the number of training samples, the time complexity of 2PKNN is higher than MCDL.\\

\begin{figure}
\begin{tabular}{cc}
\makecell{\textbf{IAPRTC-12}} &  {\textbf{ESP-GAME}}  \\
\makecell{\includegraphics[width=.46\linewidth]{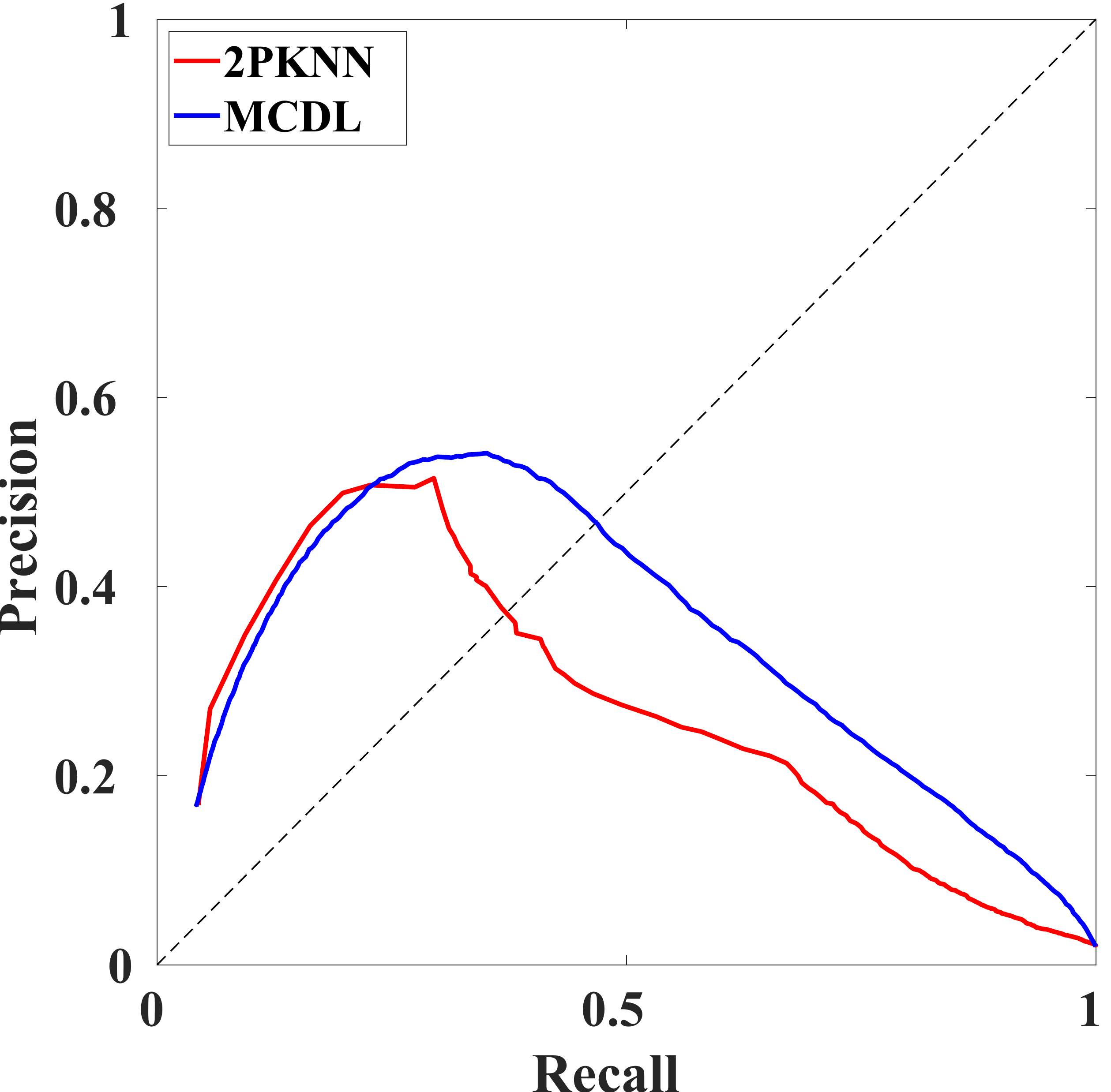}} &  \makecell{\includegraphics[width=.46\linewidth]{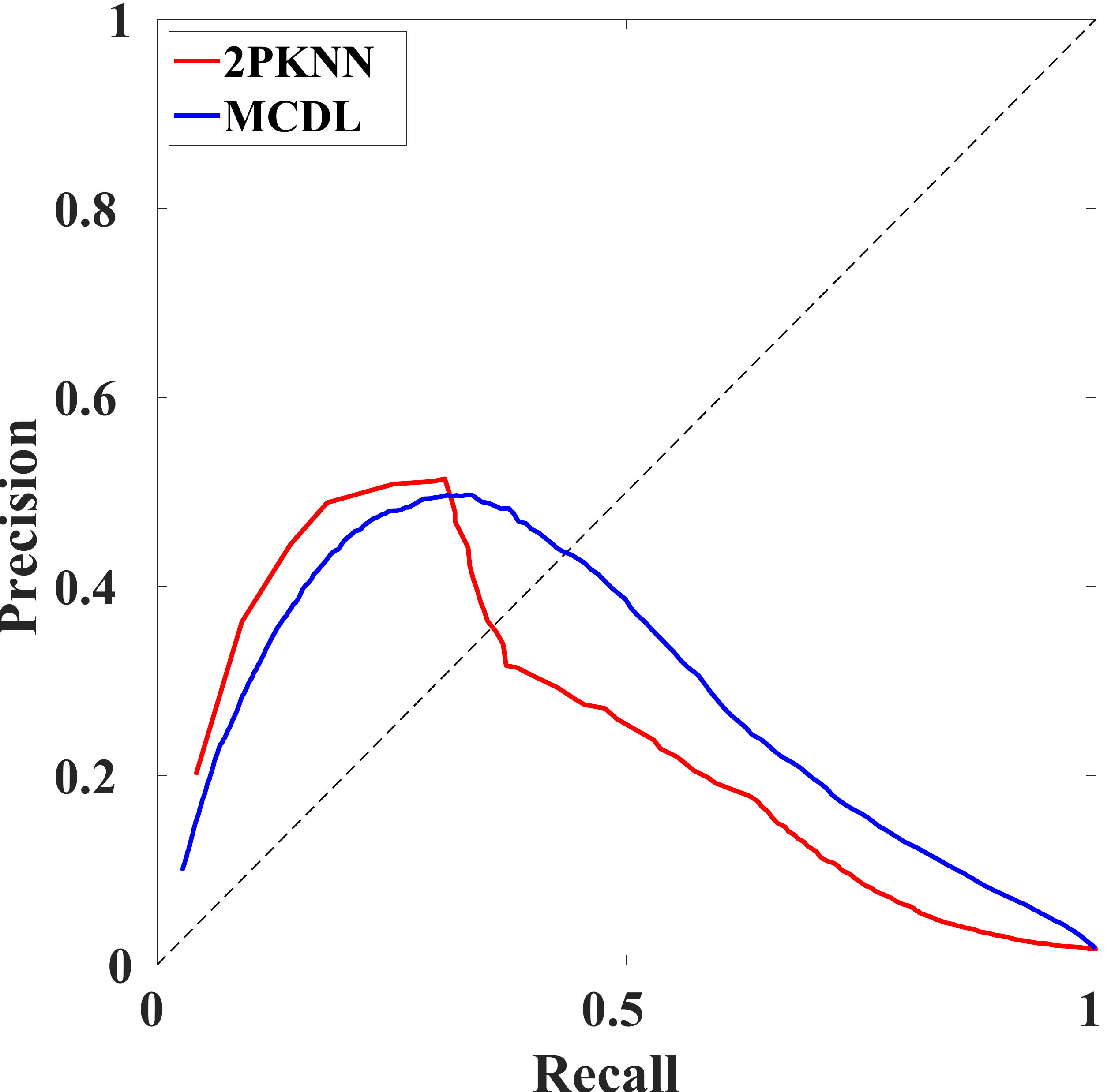}}  \\ 
\makecell{\textbf{Flickr60k}} &  {\textbf{FLICKR-125K}} \\
\makecell{\includegraphics[width=.46\linewidth]{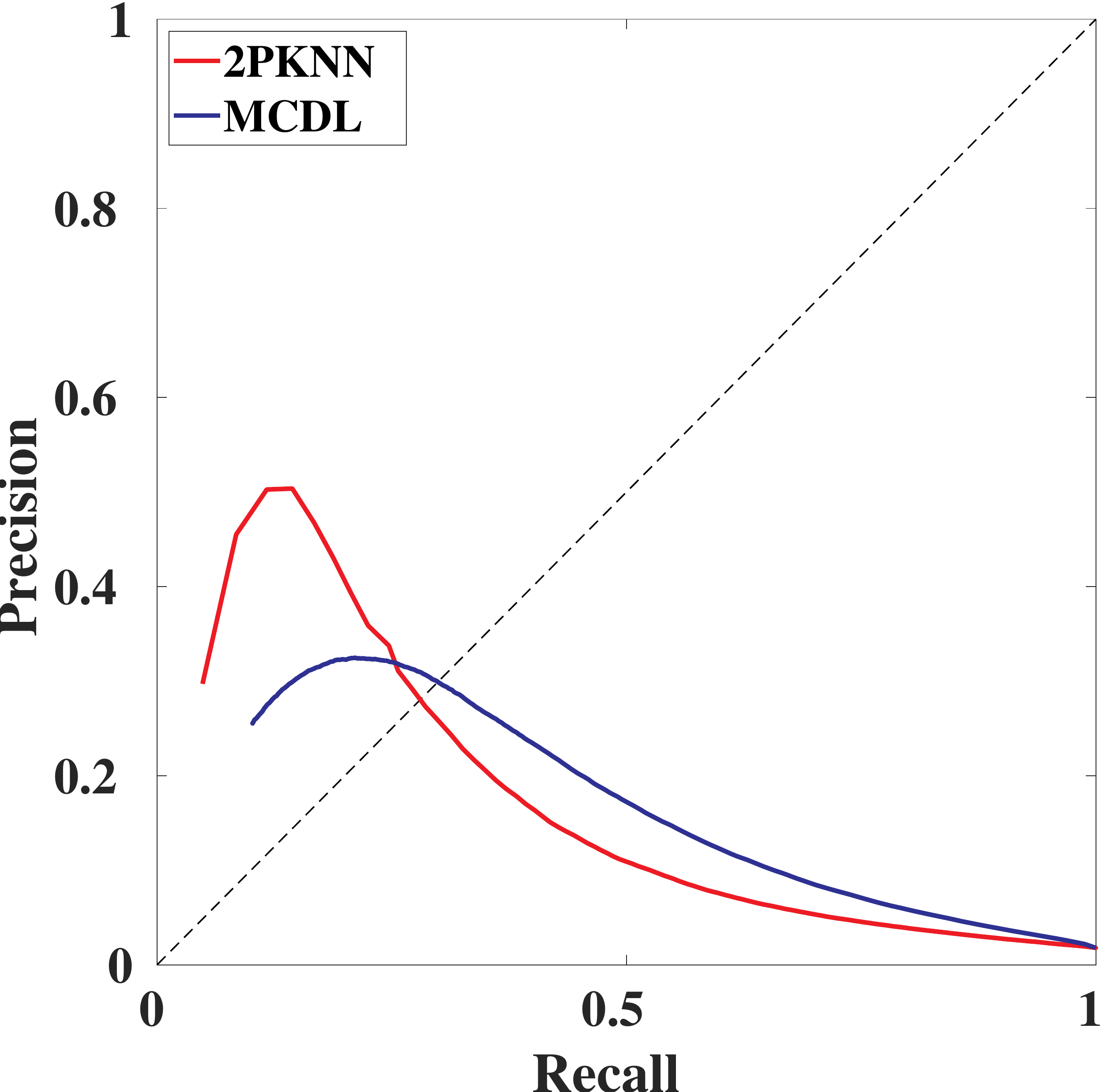}} &  \makecell{\includegraphics[width=.46\linewidth]{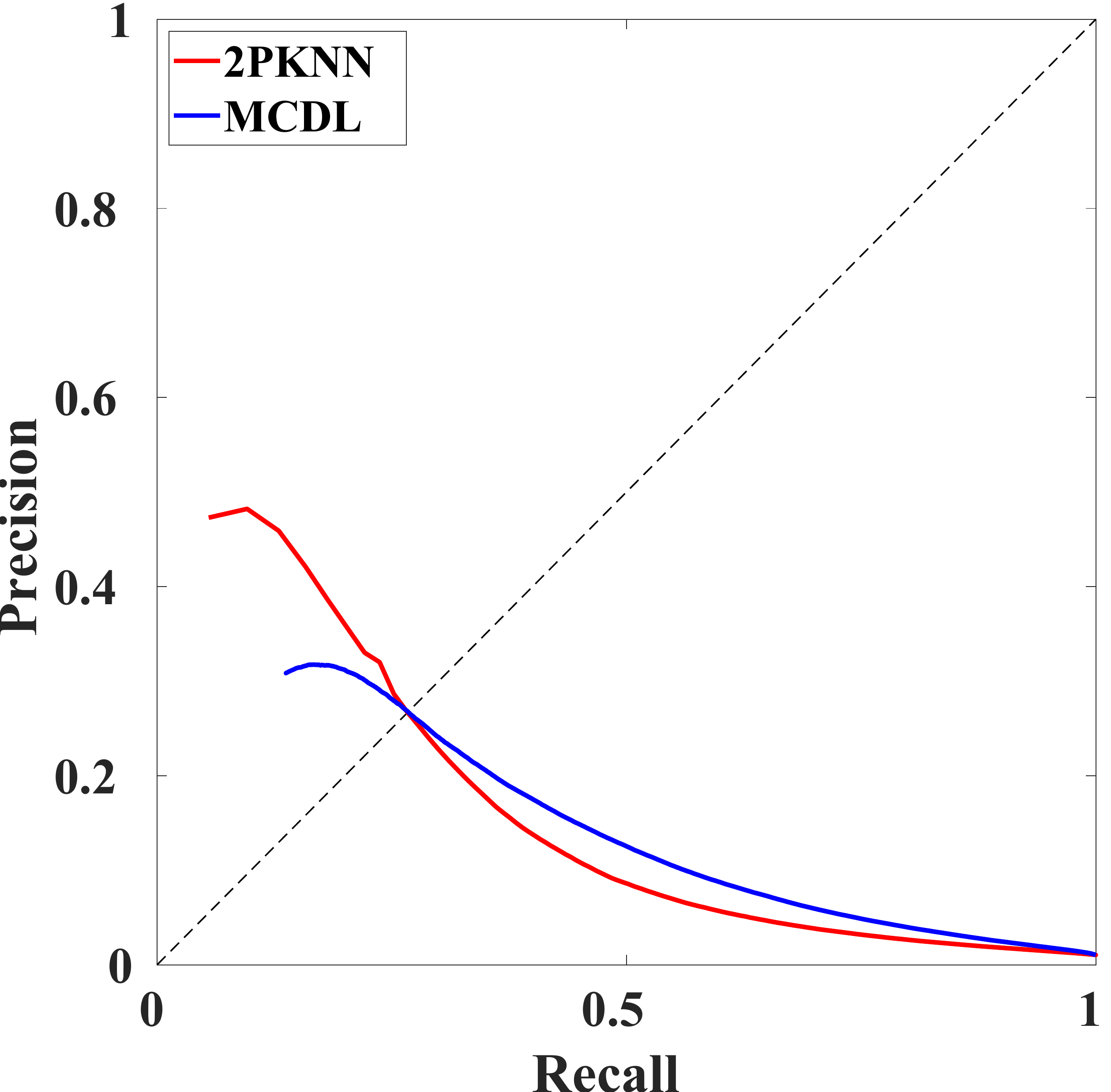}}  \\
\end{tabular}
\caption{Precision-Recall Curve for MCDL versus 2PKNN.}
\label{fig_pr_curves}
\end{figure}

\textbf{Precision-Recall\: Curves}\\
Figure \mbox{\ref{fig_pr_curves}} depicts Precision-Recall curves for our method against 2PKNN method, on four datasets. To obtain these results, we have changed the decision threshold (${\tau}_{optimal}$) and the number of assigned labels in the annotation step of MCDL and 2PKNN, respectively. The intial threshold for MCDL is 1 and the initial number of the labels for 2PKNN is also 1. So, the precision and recall do not start from one and  zero. By increasing the recall, i.e. increasing the positive labels, the precision goes through an upward trend (especially for IAPRTC-12 and ESP-GAME datasets). Then, it reaches a peak, before dropping to its minimum, where all the labels are considered positive. It stems from the fact that in contrast to the binary classification problems, in image annotation the precision and recall are computed over all the labels. Two noticable points can be concluded from the Figure \mbox{\ref{fig_pr_curves}}. The initial impression is that the area under the precision-recall curve (AUCPR) of MCDL is larger than 2PKNN on IAPRTC-12 and ESP-GAME datasets, and almost similar on the Flickr datasets. The reason is that in MCDL, we marginalize the scores, therefore by increasing the recall the number of false positives drop, which results in higher precision for MCDL versus 2PKNN. The second impression is that the optimum point of all the datasets occur for the MCDL approach which shows the superiority of this method.

\subsection{Discussion}
\label{sec_discussion}
There are some reasons that our work can outperform other existing methods. The first reason originates from prototype learning. Each prototype is a well-defined description of a set of visual features to summarize all features of a dataset. By learning these prototypes, an input image could be well-reconstructed using the minimum prototypes. The second reason is learning semantic features. For any visual prototype, its corresponding semantic part determines the association degree of any label to that specific visual prototype. The next property of our approach is controlling the weight of each visual feature in prototype learning. In fact, the more labels for a visual feature (image in dataset), the less impact on the prototypes. This stems from the fact that an image with more labels consists of less details about its labels. The fourth is about hinge loss function and its marginalized penalty. In contrast to the mean squared error loss function, it ignores the penalties out of two upper and lower bounds for positive and negative labels, respectively. Therefore, the associated labels do not impact the loss value. Lastly, L-1 regularization imposes a constraint on the semantic sparsity, which results in allocating the least and most related labels to the prototypes. All these features together would result in finding the most informative prototypes equipped with the most related labels.
%%%%%%%%%%%%%%%%%%%%%%%%%%%%%% CONCLUSION %%%%%%%%%%%%%%%%%%%%%%%%%%%%%
\section{Conclusion and Future Work}
\label{sec_conclusion}
There is a considerable redundancy in the visual and semantic contents of large-scale image datasets. MCDL provides an efficient strategy to summarize large datasets into a fewer number of prototypes with admissible accuracy. Experimental results show the superiority of the proposed method in image annotation tasks. we can reach a 90\% reduction in the annotation time while the performance is maintained or even improved in comparison to the search-based method.
MCDL method provides these benefits: Firstly, it utilizes a two-step optimization algorithm for solving a non-convex objective function which yields informative prototypes. Secondly, a marginalized loss over labels' scores is utilized to increase the generalization of the learned prototypes. Finally, the other annotation methods could leverage the prototypes extracted by MCDL. For future work, it is worth mentioning that visual features can be projected into a low-dimensionality space by embedding a transformation matrix in the optimization process, as suggested in \cite{000030-jing16}. The instances with the same labels could have more consistent sparse representation through learning such a transformation matrix. \\

%%%%%%%%%%%%%%%%%%%%%%%%%%%%%% LEMMA 1 %%%%%%%%%%%%%%%%%%%%%%%%%%%%%

\appendix
\section{Lemma 1. Convergence of Equation (\ref{eqn_lars_sc}).}
\label{sec_lemma1}
To prove the convergence of the update rule suggested in Equation (\ref{eqn_lars_sc}), it can be shown easily that $f({\alpha}_{*}^{i}) \leq g({\hat{\alpha}}^{i}[s]) \leq f({\hat{\alpha}}^{i}[s-1])$, where ${\alpha}_{*}^{i}$ is the optimum of Equation (\ref{eqn_sparsecoding}). First of all, w have (notice that $(2 y_t^i - 1) \in {\{-1,1\}}$):
\begin{equation}
\label{eqn_conv}
\begin{split}
\begin{gathered}
g({\hat{\alpha}}^{i}[s-1]) =
 {{\parallel x^i -  {\mathbf{D}}^I{{\hat{\alpha}}^{i}[s-1]} \parallel}_2^2} \\+\frac{\lambda}{{\mathcal{N}}_{+}^i} \sum_{t=1}^T {{({\tilde{y}}_t^{i}[s-1]-{\mathbf{d}}_t^L {\hat{\alpha}}^{i}[s-1])}^2}={{\parallel x^i - {\mathbf{D}}^I{{\hat{\alpha}}^{i}[s-1]} \parallel}_2^2} \: \\+ \: \frac{\lambda}{{\mathcal{N}}_{+}^i} \sum_{\{t | {\hat{\xi}}_t^{i}[s-1]>0 \}} {{({\tau}+(2 y_t^i - 1) \mathbf{C}-{\mathbf{d}}_t^L {\hat{\alpha}}^{i}[s-1])}^2} \\
 = {{\parallel x^i - {\mathbf{D}}^I{{\hat{\alpha}}^{i}[s-1]} \parallel}_2^2} \: + \: \frac{\lambda}{{\mathcal{N}}_{+}^i} \sum_{t=1}^T {{({\hat{\xi}}_t^{i})}^2} = f({\hat{\alpha}}^{i}[s-1]).
\end{gathered}
\end{split}
\end{equation}
\\
Equation (\ref{eqn_conv}) means that $g( {\alpha}^{i} ) \triangleq \underset{{\alpha}^{i} \rightarrow {\hat{\alpha}}^{i}[s-1]}{\mathrm{\lim}} f({\alpha}^{i})$ is almost a smooth approximation of $f({\alpha}^{i})$ in the current estimate of MCSC. Moreover, since ${\hat{\alpha}}^{i}[s]$ is supposed to be the optimum of $g({\alpha}^i)$, we have $g({\hat{\alpha}}^{i}[s]) \leq g({\hat{\alpha}}^{i}[s-1])$. Therefore, considering Equation (\ref{eqn_conv}), $g({\hat{\alpha}}^{i}[s]) \leq f({\hat{\alpha}}^{i}[s-1])$. Moreover, it is obvious that $f({\alpha}^i) \leq g({\alpha}^i)$, for all possible ${\alpha}^i$, because the squared loss used in $g({\alpha}^{i})$ is greater than or equal to hinge loss. So, we can now confirm our first proposition and minimizing the primary optimization problem of Equation (\ref{eqn_sparsecoding}).

\bibliographystyle{spmpsci}      % mathematics and physical sciences
\bibliography{allBib.bib}   % name your BibTeX data base

%\begin{thebibliography}{}
%
%% and use \bibitem to create references. Consult the Instructions
%% for authors for reference list style.
%\end{thebibliography}

\end{document}